\documentclass[letterpaper, 10 pt, conference]{ieeeconf}
\usepackage{amsmath}
\usepackage{amssymb}
\usepackage{adjustbox}
\usepackage{subcaption}
\usepackage{caption}
\usepackage{calc}
\usepackage{booktabs}
\usepackage{colortbl}
\usepackage{tabularx}
\usepackage{multirow}
\usepackage{array}
\usepackage[table]{xcolor}
\usepackage{tikz,pgfplots}
\usepackage[breaklinks=true,bookmarks=false,colorlinks=true]{hyperref}
\usepackage{amssymb}
\usepackage{pifont}
\usepackage{xspace}
\usepackage{cite}
\usepackage{graphicx}
\usepackage{todonotes}
\usepackage{colortbl}
\let\labelindent\relax
\usepackage{enumitem}
\usepackage{multicol}
\usepackage{makecell}
\usepackage{wasysym}
\usepackage{balance}

\renewcommand{\theequation}{\roman{equation}}

\usetikzlibrary{arrows.meta, positioning, calc, shadows}

\newcommand{\parag}[1]{\vskip2pt \noindent \textbf{#1}}

\newcommand{\cmark}{\ding{51}}
\newcommand{\xmark}{\ding{55}}
\definecolor{arrowgray}{RGB}{150,150,150}
\definecolor{profgreen}{RGB}{0,128,0}
\definecolor{darkgreen}{RGB}{0,255,0}
\definecolor{brightpurple}{RGB}{220,110,200}

\definecolor{m_red}{RGB}{255,209,209}
\definecolor{m_red_border}{RGB}{215,23,20}

\definecolor{m_orange}{RGB}{255, 215, 196}
\definecolor{m_orange_border}{RGB}{242,114,0}

\definecolor{m_green}{RGB}{197, 235, 170}
\definecolor{m_green_border}{RGB}{29,177,0}

\definecolor{m_blue}{RGB}{217,232,251}
\definecolor{m_blue_border}{RGB}{107,141,190}

\definecolor{m_yellow}{RGB}{255,242,205}
\definecolor{m_yellow_border}{RGB}{213,182,82}

\definecolor{m_gray}{RGB}{245,245,245}
\definecolor{m_gray_border}{RGB}{102,102,102}

\DeclareRobustCommand{\colorsquare}[1]{\tikz{\path[draw=#1_border,fill=#1, thick, rounded corners=0.6pt] (0,0) rectangle (6pt,6pt);}}

\def\eg{\emph{e.g.}\@\xspace} 
\def\ie{\emph{i.e.}\@\xspace}

\IEEEoverridecommandlockouts
\title{\LARGE \bf
Interactive4D: Interactive 4D LiDAR Segmentation
\vspace{-0.5cm}
}

\author{
Ilya Fradlin$^{1,\textcolor{red}{\dag}}$,
Idil Esen Zulfikar$^{1}$,
Kadir Yilmaz$^{1}$,
Theodora Kontogianni$^{2,\textcolor{red}{\mathsection}}$,
Bastian Leibe$^{1}$ \\
\small{\texttt{\url{https://vision.rwth-aachen.de/Interactive4D}}}\vspace{-1.5em}
\thanks{
$^{1}$ Computer Vision Group, RWTH Aachen University, Germany.}
\thanks{
$^{2}$ ETH AI Center, Z\"{u}rich, Switzerland.}
\thanks{
$^{\textcolor{red}{\dag}}$ This work is part of the first author’s master thesis.}
\thanks{
$\textcolor{red}{\mathsection}$ Work done at ETH AI Center. Currently at Technical University of Denmark.}
}

\newcommand{\ours}{Interactive4D}
\newcommand{\agilethreed}{AGILE3D}

\newcommand{\semantickitti}{SemanticKITTI}
\newcommand{\nuscenes}{nuScenes}
\newcommand{\kitti}{KITTI-360}
\newcommand{\lidar}{LiDAR}

\newcommand{\indomain}{in-distribution}
\newcommand{\outdomain}{zero-shot}

\begin{document}

\maketitle

\thispagestyle{empty}
\pagestyle{empty}

\begin{abstract}

Interactive segmentation has an important role in facilitating the annotation process of future \lidar{} datasets.
Existing approaches sequentially segment individual objects at each \lidar{} scan, repeating the process throughout the entire sequence, which is redundant and ineffective. In this work, we propose interactive 4D segmentation, a new paradigm that allows segmenting multiple objects on multiple \lidar{} scans simultaneously, and \ours{}, the first interactive 4D segmentation model that segments multiple objects on superimposed consecutive \lidar{} scans in a single iteration by utilizing the sequential nature of \lidar{} data. While performing interactive segmentation, our model leverages the entire space-time volume, leading to more efficient segmentation. Operating on the 4D volume, it directly provides consistent instance IDs over time and also simplifies tracking annotations. Moreover, we show that click simulations are crucial for successful model training on \lidar{} point clouds. To this end, we design a click simulation strategy that is better suited for the
characteristics of \lidar{} data. To demonstrate its accuracy and effectiveness, we evaluate \ours{} on multiple \lidar{} datasets, where \ours{} achieves a new state-of-the-art by a large margin. 
We publicly release the code and models at \texttt{\url{https://vision.rwth-aachen.de/Interactive4D}}.
\end{abstract}

\section{INTRODUCTION}

The impressive development of deep learning methods has largely been driven by the availability of large-scale annotated datasets~\cite{gupta2019lvis, kuznetsova2020open, kirillov2023segment, dai2017scannet, Hou_2019, rozenberszki2022language, behley2019semantickitti, caesar2020nuscenes, Sun2019ScalabilityIP}, particularly in the 2D domain~\cite{gupta2019lvis, kuznetsova2020open, kirillov2023segment}. However, annotating large-scale 3D datasets~\cite{dai2017scannet,Hou_2019,rozenberszki2022language,behley2019semantickitti, caesar2020nuscenes, Sun2019ScalabilityIP} remains challenging mainly due to the vast size of point clouds and the significant manual human effort required~\cite{behley2019semantickitti,dai2017scannet}.
As a result, annotated 3D datasets are scarce, impeding the development of robust 3D models. This underscores the need for efficient annotation methods tailored for 3D data. Interactive segmentation offers a promising solution to this by enabling users to create high-quality annotations with minimal effort. In this approach, the user guides the model to densely label each point in a point cloud through sparse user interactions.

This has spurred research into 3D interactive segmentation~\cite{Kontogianni2023InterObject3D, Yue2024AGILE3D, Sun2023ACI, Han2024ClickFormer}. Early efforts primarily focused on indoor point clouds, with initial work~\cite{Kontogianni2023InterObject3D} framing the task as single-object interactive segmentation.
In this setting, annotators segment each object individually by providing positive clicks on the object and negative clicks on other areas, essentially treating it as a binary segmentation problem. Recently,~\cite{Yue2024AGILE3D} reformulated the task as multi-object interactive segmentation, where annotators segment multiple objects simultaneously. Here, the positive clicks for one object inherently serve as negative clicks for other objects, better utilizing user input and increasing efficiency. Both methods only consider interactively segmenting object instances, \ie, \textit{things}, while neglecting amorphous regions, \ie, \textit{stuff}.

\begin{figure}[t!]
\centering
\includegraphics[width=\linewidth]{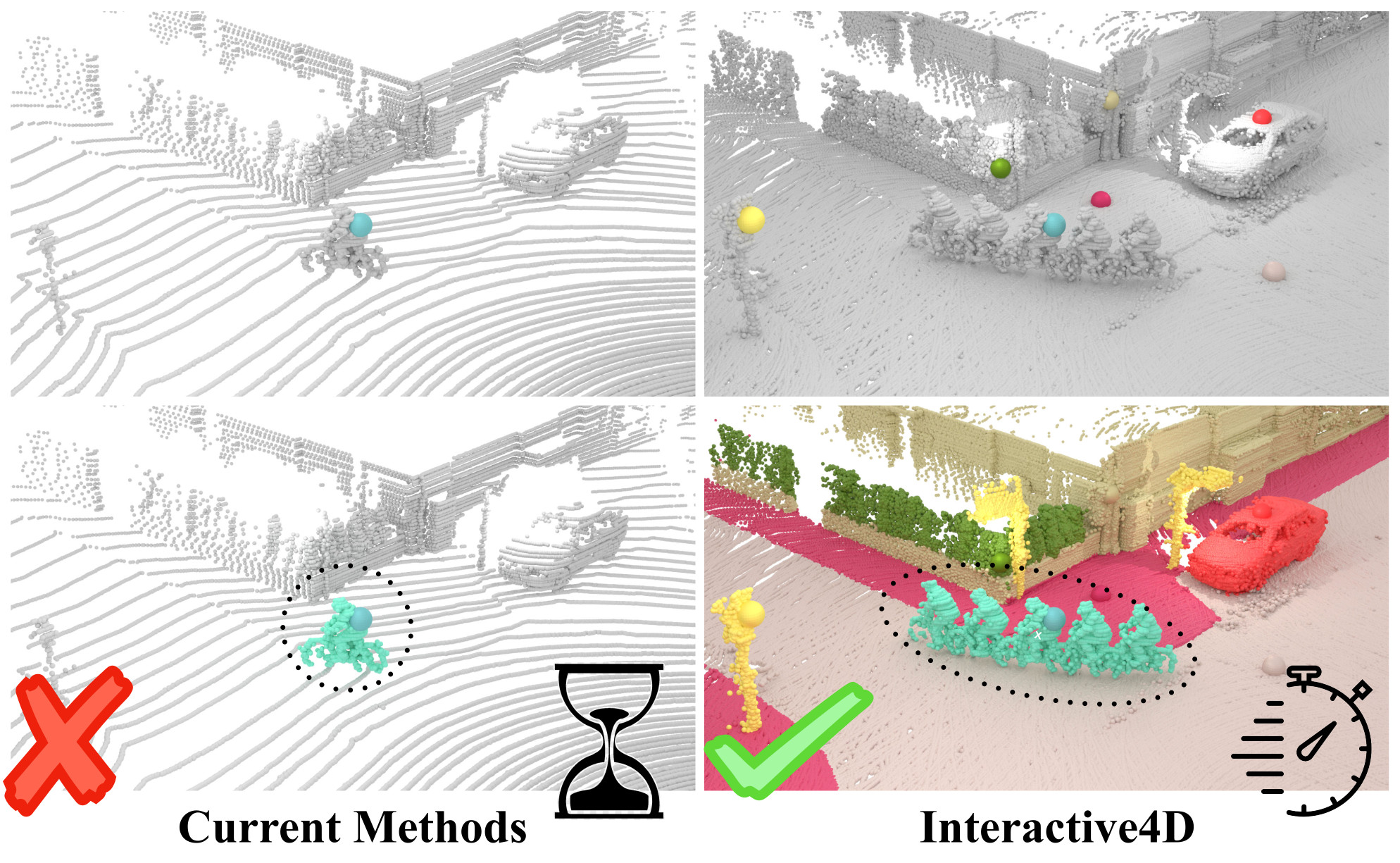}
\vspace{-0.5cm}
\caption{\textbf{\textit{Left:}} Current interactive \lidar{} segmentation methods~\cite{Sun2023ACI, Han2024ClickFormer} segment each object and each \lidar{} scan individually, which is sub-optimal. 
\textbf{\textit{Right:}} In contrast, \ours{} segments multiple objects on superimposed consecutive \lidar{} scans at once, significantly improving efficiency, while providing consistent instance IDs over time.
}
\label{fig:teaser}
\end{figure}

Despite the successes on indoor point clouds, interactive segmentation of outdoor \lidar{} point clouds remains under-explored.
A few recent works~\cite{Sun2023ACI, Han2024ClickFormer} have tackled this task,
yet they follow the single-object paradigm, which is shown to be less efficient than the multi-object paradigm.
Also, they treat each \lidar{} scan as an independent entity, ignoring the sequential nature of \lidar{} scans (see Fig.\,\ref{fig:teaser}, \emph{left}). Given that \lidar{} sensors operate at high frequencies, successive scans capture overlapping regions. Thus, independently annotating each scan is inefficient, leading to unnecessary annotation effort. 
Furthermore, independently annotating each scan complicates the task of maintaining consistent instance IDs across consecutive scans, which is essential for tracking tasks.

Having realized these limitations, in this work, we apply \textit{segmenting everything all at once} strategy for \lidar{} data and propose interactive 4D segmentation, a new paradigm where the annotator segments \textit{multiple} objects on \textit{multiple} \lidar{} scans simultaneously. To show the effectiveness of this paradigm, we propose \ours{}, the first interactive 4D segmentation model that performs multi-object segmentation on superimposed consecutive \lidar{} scans for both \textit{things} and \textit{stuff} objects. This improves efficiency by enabling multi-object interactive segmentation across the entire 4D space-time \lidar{} volume. By working directly on 4D data, \ours{} inherently ensures consistent instance IDs on superimposed consecutive \lidar{} scans (see Fig.\,\ref{fig:teaser}, \emph{right}). This makes it highly adaptable for tracking tasks, simplifying the annotation process for \lidar{} tracking datasets, while also paving the way for future research directions.

In the interactive community~\cite{mahadevan2023itis,Kontogianni2023InterObject3D, Yue2024AGILE3D}, simulated clicks are commonly used for both training and testing. Many methods typically mimic a user
who always clicks at the center of the largest error region~\cite{mahadevan2023itis, Yue2024AGILE3D,kontogianni2020continuous}, while others select clicks randomly~\cite{Kontogianni2023InterObject3D,li2018interactive,Sun2023ACI,Han2024ClickFormer}. Both approaches have notable limitations with LiDAR data. The former incurs high computational costs, resulting slower run-times, while both approaches struggle with the sparsity and size variations between small and large objects—common issues in outdoor scenes—that often lead to poor segmentation quality.
To address these limitations, we propose a new click simulation strategy for both training and evaluation. It generates enhanced and scale-invariant click simulations by accounting for the sparse nature of \lidar{} point clouds, identifying the most relevant areas for clicking, and effectively managing scale variations between small and large objects.

\begin{figure*}[t!]
\centering
\includegraphics[width=1.0\textwidth]{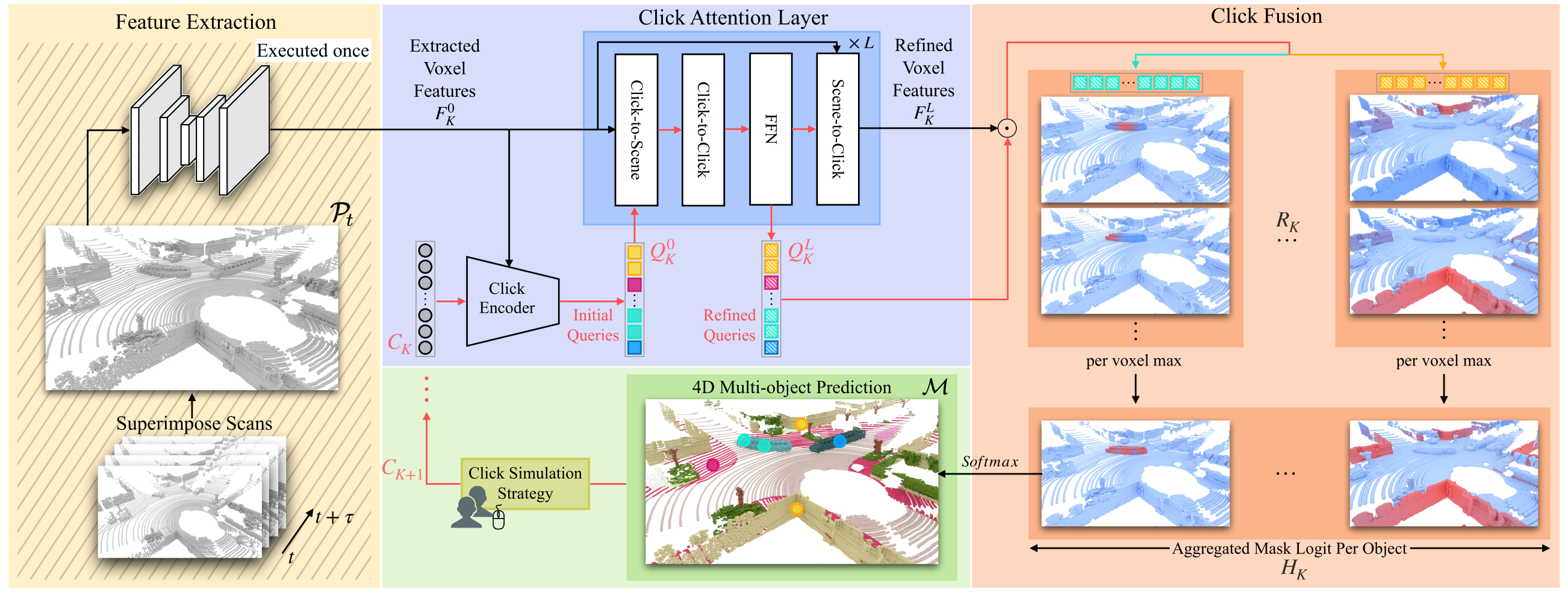}
\caption{\textbf{Overview.} \colorsquare{m_yellow}: We superimpose consecutive scans into a single point cloud and extract per-voxel features (executed once). \colorsquare{m_blue}: The clicks are encoded as initial queries, then refined through multiple attention layers. \colorsquare{m_orange}: The dot product between refined queries and voxel features results in click responses, which are fused in click fusion module to form predictions \colorsquare{m_green}.}
\label{fig:model}
\end{figure*}

Following the evaluation protocol from~\cite{
Rana2023DynaMITe, Kontogianni2023InterObject3D, Yue2024AGILE3D
}, we train Interactive4D on SemanticKITTI~\cite{behley2019semantickitti} and evaluate in multiple datasets to assess both in-distribution and zero-shot performance. We report results on the single-object, multi-object and 4D interactive segmentation setups, achieving state-of-the-art performance across all setups. To evaluate real-world generalization, we integrated Interactive4D into a user interface~\cite{Yue2024AGILE3D} and conducted a user study where participants annotated selected scenes. The study showed that Interactive4D performs well not only with simulated clicks but also in real use cases.

In summary, \textbf{our contributions are:} \textbf{(1)} We propose interactive 4D segmentation, a new paradigm that encompasses interactive segmentation of \textit{multiple} objects on \textit{multiple} scans at once by leveraging the sequential nature of \lidar{} data. 
\textbf{(2)} We introduce \ours{}, the first interactive 4D segmentation model capable of accurately segmenting both \textit{things} and \textit{stuff} on spatio-temporal point clouds, while also providing consistent instance IDs over time for tracking tasks.
\textbf{(3)} We design a novel click simulation strategy that is more suitable for the characteristics of \lidar{} data.
\textbf{(4)} We achieve state-of-the-art performance by a significant margin on several \lidar{} datasets
and prove the effectiveness of \ours{} in real annotation cases through a user-study with human annotators.

\section{RELATED WORK}
\parag{\lidar{} Panoptic Segmentation and Tracking.} 
\lidar{} Panoptic Segmentation (LPS)~\cite{9340837,behley2021benchmark,fong2022panoptic} unifies semantic and instance segmentation of \lidar{} point clouds.
Recently, it has been extended to sub-task tracking with 4D Lidar Panoptic Segmentation (4D-LPS)~\cite{aygun20214d}, which performs semantic, instance segmentation, and tracking jointly. Both LPS~\cite{sirohi2021efficientlps, hurtado2020mopt, razani2021gp, zhou2021panoptic, hong2021lidar, li2022panoptic, su2023pups, marcuzzi2023mask} and 4D-LPS~\cite{aygun20214d, marcuzzi2022contrastive, kreuzberg20224d, Hong2022LiDARbased4P, agarwalla2023lidar, zhu20234d,athar20234dformer,marcuzzi2023mask4d, Yilmaz2024Mask4Former} methods follow similar algorithmic paradigms.
The fundamental difference between them is that LPS methods operate on single \lidar{} scans, while 4D-LPS methods mainly operate on superimposed consecutive \lidar{} scans to accomplish tracking. Depending on whether it operates on a single scan or superimposed consecutive scans, \ours{} can function as either an LPS or 4D-LPS method, assuming the user provides semantic labels of the predicted masks. It performs segmentation and tracking by incorporating user inputs, capable of improving the results with refinement clicks. Later, we show that \ours{} outperforms state-of-the-art results both in LPS and 4D-LPS tasks with minimal user input, and further improves upon them with additional user input.

\parag{Interactive 3D Segmentation.} 2D interactive segmentation is well established~\cite{kirillov2023segment}, however, adopting it to generate 3D labels leads imperfections due to differences in field-of-view, perspective, and calibration errors~\cite{le2024jrdb}. To this end, InterObject3D~\cite{Kontogianni2023InterObject3D} tackled the interactive segmentation for indoor point clouds, focusing on single-object interactive segmentation. Later, AGILE3D~\cite{Yue2024AGILE3D} proposed multi-object interactive segmentation for indoor point clouds, enhancing efficiency significantly. Inspired by ~\cite{Yue2024AGILE3D}, we explore multi-object interactive segmentation for \lidar{} point clouds and also go one step further by extending multi-object \lidar{} interactive segmentation to the 4D setup, aiming to maximize efficiency. A few recent works~\cite{Sun2023ACI, Han2024ClickFormer} 
have studied 3D interactive segmentation for outdoor \lidar{} point clouds.
CRSNet~\cite{Sun2023ACI} focuses on interactively segmenting only \textit{things} objects and follows the single-object paradigm in LiDAR data.
ClickFormer~\cite{Han2024ClickFormer}, a concurrent work, interactively segments both \textit{things} and \textit{stuff}, and addresses the scale disparity of objects in \lidar{} data by populating extra augmentation clicks across the scan, again following single-object paradigm.
In contrast, \ours{} is designed to handle multi-object interactive \lidar{} segmentation of both \textit{things} and \textit{stuff} in the 4D setup,  maximizing efficiency by leveraging the context holistically both in space and time.

\section{METHOD}
\label{sec:method}

Inspired by the success of attention-based models for interactive segmentation~\cite{Yue2024AGILE3D, Rana2023DynaMITe}, we add key technical modifications to leverage the full potential of such models in \lidar{} point clouds and introduce \ours{}, our interactive 4D segmentation model as depicted in Fig.~\ref{fig:model}. For clarity, we present the entire process using matrix notation.

\parag{Spatio-Temporal Point Cloud.} (Fig.~\ref{fig:model}, \colorsquare{m_yellow}) 
We begin with superimposing consecutive \lidar{} scans within a short temporal window $[t, t+\tau]$ into a single spatio-temporal point cloud $\mathcal{P}_t$~$\in$~$\mathbb{R}^{M\times3}$. This representation is beneficial for interactive segmentation as: \textbf{(1)} Static objects remain in the same spatial region across scans, and annotating them becomes more efficient, requiring fewer clicks to achieve the desired accuracy. \textbf{(2)} Dynamic objects, on the other hand, appear as multiple silhouettes, reflecting their movement over time and enabling intuitive tracking by associating silhouettes within a single point cloud. \textbf{(3)} The unified point cloud also offers higher point density compared to individual scans, making objects more concentrated and easier to recognize. This is particularly beneficial for identifying smaller objects, which are often difficult to detect in sparse \lidar{} data.

\parag{Feature Extractor.}
We voxelize $\mathcal{P}_t$, resulting in $\mathcal{V}_t$~$\in$~$\mathbb{Z}^{N \times 3}$ to enable efficient processing using 3D sparse convolutions which operate over a grid.
Time is included as an additional feature in $\mathcal{V}_t$ to distinguish between voxels from different \lidar{} scans.
To extract per-voxel features $\mathcal{F}^0$~$\in$~$\mathbb{R}^{N \times D}$, we use a 3D sparse U-Net~\cite{choy20194d} in line with~\cite{robert2022learning, xie2020pointcontrast, huang2021predator,Yue2024AGILE3D}.
\parag{Click Encoder.}
(Fig.~\ref{fig:model}, \colorsquare{m_blue}) Given a set of raw clicks $C_K$ for the $K$-th iteration, the goal of the click encoder is to encode $C_K$ as click queries $Q^0_K$~$\in$~$\mathbb{R}^{K \times D}$.
The initial queries $Q^0_K$ serve as a starting point for the refinement and should capture the relevant information to effectively represent the regions the user aims to segment.
We formulate it as: 
\begin{equation}
    Q^0_K = \mathcal{E}_f+ \mathcal{E}_{xyz,t} + \mathcal{E}_k  + \mathcal{E}_{id}
\end{equation}
where $\mathcal{E}_f$ and $\mathcal{E}_{xyz,t}$ are the extracted features and the positional encoding~\cite{tancik2020fourfeat} of the clicked voxel respectively. $\mathcal{E}_k$ are the iteration encodings~\cite{vaswani2017attention} representing the ordering of the clicks.
Additionally, unlike prior works~\cite{Yue2024AGILE3D, Han2024ClickFormer}, we explicitly encode the associated object IDs through a separate learned embedding $\mathcal{E}_{id}$.
This allows clicks associated with the same object to be identified as related and distinguished from other clicks during the refinement process.
\parag{Refinement.} (Fig.~\ref{fig:model}, \colorsquare{m_blue}) This module consists of $L$ consecutive click attention layers that refine both the click queries $Q^0_K$ and the voxel features $\mathcal{F}_K^0$.
In each layer, $Q^l_K$ attend to $\mathcal{F}_K^l$ through cross-attention.
Then, $Q^l_k$ self-attend to each other.
Finally, $\mathcal{F}_K^l$ cross attend to $Q^l_K$ to refine feature representations.
This progressive refinement is repeated across $L$  layers, resulting in final $Q^L_K$ and $\mathcal{F}^L_K$.

\parag{Click Fusion.} (Fig.~\ref{fig:model}, \colorsquare{m_orange}) After the final refinement, the dot product between $Q^L_K$ and $\mathcal{F}^L_K$ results in click response maps $R_K = Q^L_K \cdot (\mathcal{F}^L_K)^T$~$\in$~$\mathbb{R}^{K \times N}$  representing the response of each voxel to each click.
To generate object-level heatmaps $H_K \in \mathbb{R}^{I\!D \times N}$, we apply a per-voxel maximum operation across all click responses associated with the same object.
This ensures that each click contributes only to the region where it has the highest response, resulting in aggregated object heatmaps.
Then, the final mask  $\mathcal{M} \in \mathbb{R}^N$ is obtained by applying $\mathit{Softmax}$ over the ID dimension of $H_K$.
\parag{Localized Loss.} (Fig.~\ref{fig:model}, \colorsquare{m_orange}) To train the model, we use a combination of the cross-entropy and the dice loss~\cite{milletari2016fully}.
\begin{equation}
\mathcal{L} = \frac{1}{N} \sum\nolimits_{p \in P} w_p \left( \lambda_{\text{CE}} \cdot \mathcal{L}_{\text{CE}}(p) + \lambda_{\text{Dice}} \cdot \mathcal{L}_{\text{Dice}}(p) \right)    
\end{equation}

Here, $\lambda_{\text{CE}}$ and $\lambda_{\text{Dice}}$ are scalars balancing the two losses.
The weight factor $w_p$ adjusts the loss based on the proximity of each point to user clicks, making the loss more localized around each click.
It is formulated as follows:
\begin{equation}
\begin{cases}
w_{\max} - (w_{\max} - w_{\min}) \cdot \tilde{d}_p, & \text{for } 0 \leq \tilde{d}_p \leq 1 \\
w_{\min}, & \text{otherwise}
\end{cases}    
\end{equation}

where $\tilde{d}_p$ is the normalized distance between point $p$ and its nearest user click, scaled by $\delta$.
This formulation ensures that points within $\delta$ meters of a click receive weights that decrease linearly from $w_{\max}$ to $w_{\min}$ as the distance increases, making the loss more localized.
This loss design along with the click fusion operation effectively forces each click to be more local, giving a strong response around the clicked region, ensuring each click adds information without interfering with others.
At the same time, points further than $\delta$ get a weight of $w_{\min}$ providing some incentive for clicks to segment the far away parts of the object, needed to effectively handle large and easy-to-segment regions such as road.

\parag{4D Inference.} (Fig.~\ref{fig:model}, \colorsquare{m_green}) Within each short temporal window $[t, t + \tau]$ we directly obtain consistent instance IDs by assigning each point to the object with the highest response in $H_K$.
However, tracking tasks require consistent instance IDs over the entire sequence. 
To achieve this, we form temporal windows with one overlapping \lidar{} scan~\cite{aygun20214d} and use the predictions of both temporal windows in this scan to carry instance IDs from $\mathcal{P}_t$ to $\mathcal{P}_{t+\tau}$.
This approach also enables us to parallelize the annotation process among multiple annotators while automatically ensuring consistent instance IDs across the entire sequence.

\begin{figure}[t]
\centering
\includegraphics[width=1\linewidth]
{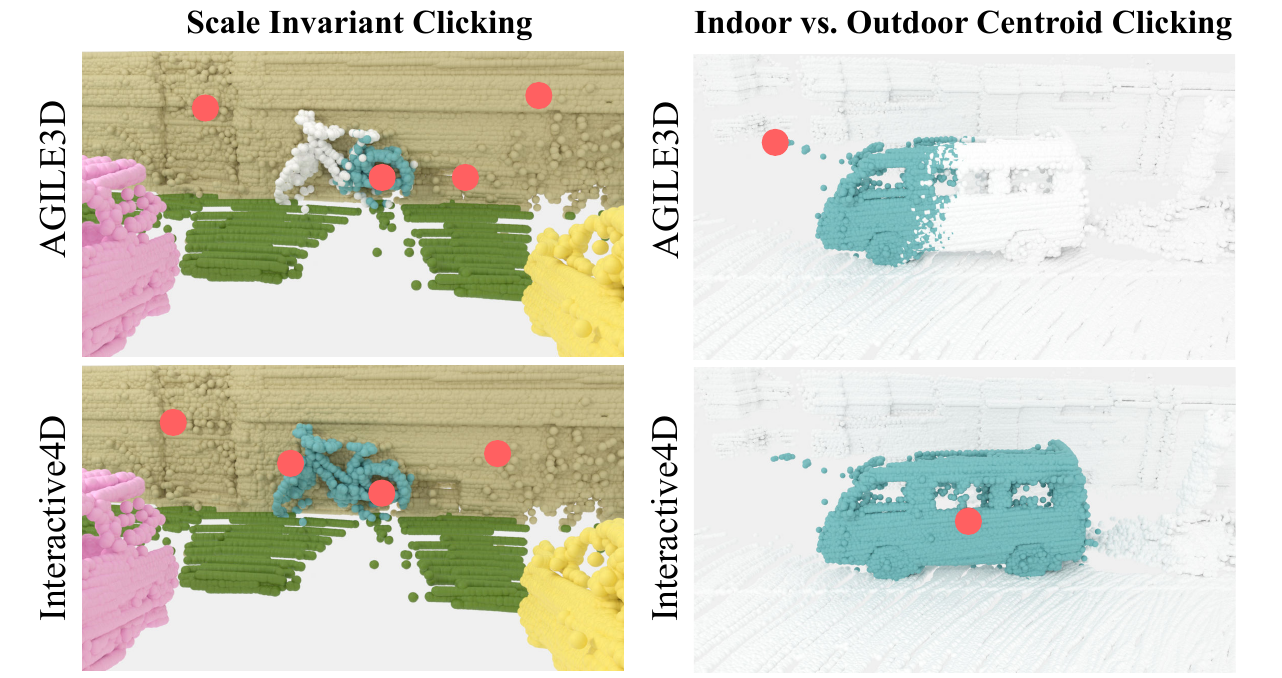}
\vspace{-0.5cm}
\caption{\small{Examples for centroid clicking and scale-invariant clicking.}}
\label{fig:clicking demonstration}
\end{figure}

\parag{Click Simulation Strategy.}\label{sec:click_sim} (Fig.~\ref{fig:model}, \colorsquare{m_green}) Interactive segmentation models rely on annotator input to iteratively refine predictions, yet involving humans during training is impractical. Instead, synthetic clicks are simulated based on predictions and ground truth. 
Simulation strategies should: \textbf{(1)} focus the model's learning on error regions to improve accuracy with fewer interactions, and \textbf{(2)} minimize the gap between training and real-world usage.
Two main types of click simulations are employed in interactive 3D segmentation models~\cite{Kontogianni2023InterObject3D, Yue2024AGILE3D, Sun2023ACI, Han2024ClickFormer}. 
Inspired by 2D simulations~\cite{xu2016deep, li2018interactive, mahadevan2023itis, kontogianni2020continuous, sofiiuk2022reviving, chen2022focalclick} models operating under dense data ~\cite{Kontogianni2023InterObject3D, Yue2024AGILE3D} employ a Boundary Dependent (BD) click strategy. This approach selects the point furthest from the boundary by using the following metric: 
\begin{equation} \label{eq:former_clicking}
    p_{\text{click}_{i\rightarrow j}} = \arg \max\nolimits_{p \in \mathit{E}_{i\rightarrow j}} \left( \min\nolimits_{q \in \mathcal{P}_t \setminus \mathit{E}_{i\rightarrow j}} \|q - p\|^2 \right)
\end{equation}
Here,  $E_{i\rightarrow j}$ is the error region consisting of points misclassified as object $j$ on the ground truth object $i$, $p$, and $q$ are the points within and outside the error region respectively. While \textit{BD} is effective for single object setup and dense data, under multi-object-\lidar{} setup, it becomes computationally intensive due to its pairwise distance calculations which are time-consuming and memory demanding. More recent methods~\cite{Sun2023ACI, Han2024ClickFormer} mitigate this issue by adopting fully random clicking, significantly reducing the required computation. However, both methods still suffer from two key limitations: \textbf{(1) Bias Towards Larger Objects}: In multi-object interactive segmentation, the error region must first be identified since errors can exist across various objects. \textit{BD} selection implicitly determines the error size of the region by switching the $\arg \max$ in Eq.~\eqref{eq:former_clicking} with $\max$ operation. This approach tends to bias clicks toward larger objects (e.g., buildings) while overlooking smaller ones (e.g., bicycles) (see Fig.~\ref{fig:clicking demonstration}, \emph{left}). The same issue applies to the random click strategy, as the random distribution naturally overlooks smaller objects, making them underrepresented during training. \textbf{(2) Non-informative Initial Clicks}: In dense data, selecting the point farthest from the boundary is effective, as it often captures the ``center'' of an error region. However, in sparse \lidar{} point clouds, this often leads to misplaced clicks near the periphery of the error region due to surrounding empty space (see Fig.~\ref{fig:clicking demonstration}, \emph{right}). This occurs because points outside the error region determine the boundary. Random clicking exclusively faces a similar issue, as the selection lacks focus on a specific error region and instead targets the entire set of errors indiscriminately.

Aiming to solve the mentioned drawbacks, we propose a new click selection strategy, dividing the process into two separate steps: \textbf{(1) Scale Invariant Error Region Selection (SI):}
To counter the over-prioritization of larger objects, we propose an IoU-based metric for determining the largest error region, ensuring scale invariance:
\begin{equation}\label{eq: si clicking}
    \mathit{S}(\mathit{E}_{i\rightarrow j}) = \left(|\mathit{E}_{i\rightarrow j}| \cdot {|\mathit{GT}_i|}^{-1} \right)\cdot {\mathit{IoU}_i}^{-1}
\end{equation}
This metric, where $ | \cdot |$ indicates the number of points, and $GT_i$ is the ground truth object, balances the proportion of the object belonging to the error region, and the current segmentation accuracy of the ground truth object, indicating potential improvement.
This approach prioritizes error regions covering a substantial portion of the object, particularly those with low IoU, preventing smaller objects from being overlooked. This results in a more balanced distribution of clicks during training, allowing the model to learn more comprehensive feature representations. \textbf{(2) Enhanced Click Selection within Error Regions:} Similar to prior works that rely on \textit{BD}, we aim to select the ``center'' of the ground truth mask for the initial click, as this often captures the key characteristics of the region. 
To ensure informative selection in sparse domains, we define the center as the point closest to the object’s centroid, typically capturing the most representative area for a solid segmentation start. 
However, for refinement clicks—where error regions become small and fragmented across the point cloud—the effectiveness of centroid selection decreases and is challenging for real users to follow. To address this, we propose switching to random point selection within the error region for refinement clicks. Unlike~\cite{Sun2023ACI, Han2024ClickFormer}, we only perform the random selection after the error region has been identified.
The injection of randomness makes the model more robust to user behavior\,(see Sec.\,\ref{sec:click_randomness}) by learning diverse features. 

Our approach simplifies and decouples the decision process.
It accelerates training and reduces memory requirements significantly since the click simulator is called multiple times during training.  
Unlike methods that either limit the number of objects  ~\cite{Han2024ClickFormer, Yue2024AGILE3D} or concentrate only on \textit{things}~\cite{Sun2023ACI} to manage computational costs, our method avoids these trade-offs. This enables faster training, holistic segmentation, and scalability to 4D data, where the number of points increases significantly.
Fig.~\ref{fig:clicking demonstration} shows the effects of our proposed strategy modifications overall.

\vspace{-0.5em}

\section{EXPERIMENTS}
\vspace{-0.5em}
\parag{Datasets.} Interactive segmentation involves two key scenarios: (1) Training the model on the annotated portion of the target dataset to assist in labeling the remaining data, referred to as \textit{\indomain{}}. (2) Training the interactive model on a pre-annotated dataset and then applying it to annotate a different target dataset, known as \textit{\outdomain{}}. The first scenario highlights its effectiveness in speeding up annotation when partial labels are already available, while the second scenario evaluates the model's ability to generalize to new, unseen data distributions. 
To this end, we use two well-established public \lidar{} datasets, \semantickitti{}~\cite{behley2019semantickitti} and \nuscenes{}~\cite{caesar2020nuscenes}.
For all experiments, we train \ours{} on the training split of \semantickitti{}. The \textit{\indomain{}} setup is evaluated using the validation split of \semantickitti{}, while the \textit{zero-shot} performance is evaluated on the validation split of \nuscenes{}.

\newcolumntype{g}{>{\columncolor{red!20}}c}
\begin{table}[t]
\renewcommand{\arraystretch}{1.2}
\center
\tabcolsep=0.06cm
\footnotesize{
\caption{\small\textbf{In-distribution Interactive \lidar{} Segmentation.}}
\label{table:within_domain}
\begin{tabular}{cl c cccccc c ccc}
\toprule
\multicolumn{13}{c}{\textbf{Training}: \semantickitti{}  $\rightarrow$ \textbf{Evaluation}: \semantickitti{}} \\
\midrule
\multicolumn{2}{c}{} & & \multicolumn{6}{c}{$\overline{\text{IoU}}\text{@}\overline{\text{k}}\uparrow$} & & \multicolumn{3}{c}{$\overline{\text{NoC}}\text{@}\overline{\text{q}}\downarrow$} \\
\cmidrule(lr){4-9} \cmidrule(lr){11-13}
& \textbf{Method} & & @1 & @2 & @3 & @4 & @5 & @10 & & @80 & @85 & @90 \\
\midrule
\multirow{2}{*}{\rotatebox{90}{Single}} & \agilethreed~\cite{Yue2024AGILE3D} & & 53.1 & 63.7 & 70.0 & 74.0 & 76.7 & 83.0 & & 3.94 & 4.40 & 4.98\\
& \cellcolor{gray!10}\textbf{\ours} & \cellcolor{gray!10} & \cellcolor{gray!10}67.5 & \cellcolor{gray!10}73.9 & \cellcolor{gray!10}78.3 & \cellcolor{gray!10}81.3 & \cellcolor{gray!10}83.4 & \cellcolor{gray!10}88.2 & \cellcolor{gray!10} & \cellcolor{gray!10}3.38 & \cellcolor{gray!10}3.85 & \cellcolor{gray!10}4.46\\
\midrule
\multirow{2}{*}{\rotatebox{90}{Multi}} & \agilethreed~\cite{Yue2024AGILE3D} & & 70.1 &  75.4 & 78.6 & 81.2 & 83.3 & 86.6 & & 2.82 & 3.29 & 3.98 \\
& \cellcolor{gray!10}\textbf{\ours} & \cellcolor{gray!10} & \cellcolor{gray!10}77.4 & \cellcolor{gray!10}84.9 & \cellcolor{gray!10}87.0 & \cellcolor{gray!10}88.3 & \cellcolor{gray!10}89.1 & \cellcolor{gray!10}91.2 & \cellcolor{gray!10} & \cellcolor{gray!10}2.53 & \cellcolor{gray!10}2.97 & \cellcolor{gray!10}3.62 \\
\midrule
\multirow{2}{*}{\rotatebox{90}{4D}} & \agilethreed~\cite{Yue2024AGILE3D} & & 81.5 & 84.3 & 84.9 & 84.9 & 85.0 & 85.1 & & 2.06 & 2.59 & 3.42 \\
& \cellcolor{gray!10}\textbf{\ours} & \cellcolor{gray!10} & \cellcolor{gray!10}\textbf{87.7} & \cellcolor{gray!10}\textbf{89.3} & \cellcolor{gray!10}\textbf{90.6} & \cellcolor{gray!10}\textbf{91.3} & \cellcolor{gray!10}\textbf{91.7} & \cellcolor{gray!10}\textbf{92.8} & \cellcolor{gray!10} & \cellcolor{gray!10}\cellcolor{gray!10}\textbf{1.26} & \cellcolor{gray!10}\textbf{1.64} & \cellcolor{gray!10}\textbf{2.28} \\
\bottomrule
\end{tabular}
}
\end{table}

\begin{table}[t]
\renewcommand{\arraystretch}{1.2}
\center
\tabcolsep=0.12cm
\footnotesize{
\caption{\textbf{\small{Non-interactive 3D \& 4D \lidar{} Panoptic Segmentation comparison on \semantickitti{}.}}
}
\label{table:3d_4d_panoptic}
\begin{tabular}{l cccc c ccc}
\toprule
\parbox[t]{1mm}
&  & \multicolumn{3}{c}{3D} & & \multicolumn{3}{c}{4D} \\
\cmidrule{3-5} \cmidrule{7-9}
\textbf{Method} & {\#}$_{{\text{clicks}}}$ & PQ & SQ & RQ & & LSTQ & S$_\text{assoc}$ & S$_\text{cls}$\\
\midrule
Mask4Former~\cite{Yilmaz2024Mask4Former} & - & 61.7 & 81.0 & 71.4 & & 70.5 & 74.3 & 66.9 \\
\arrayrulecolor{black!10}\cmidrule{1-9}\arrayrulecolor{black}
\multirow{5}{*}{\textbf{\ours}} & 1 & 70.5 & 84.0 & 82.4 & & 82.8 & 80.9 & 84.7 \\
 & 2 & 82.9 & 86.7 & 95.2 & & 84.7 & 81.7 & 87.9 \\
 & 3 & 85.9 & 88.6 & 96.7 & & 85.8 & \textbf{82.8} & 88.9 \\
 & 5 & 88.7 & 90.4 & 97.9 & & \textbf{85.9} & 82.5 & 89.3 \\
 & 10 & \textbf{91.3} & \textbf{92.2} & \textbf{99.0} & & 85.5 & 81.3 & \textbf{90.0} \\
\bottomrule
\end{tabular}
}
\end{table}

\parag{Evaluation Protocol.}
For our experiments, we use standard evaluation metrics, as outlined in previous works~\cite{Kontogianni2023InterObject3D,Yue2024AGILE3D}: (1) $\overline{\text{IoU}}@\overline{\text{k}}$ evaluates the average intersection-over-union~(IoU) achieved with k clicks per object, averaged across all objects. (2) $\overline{\text{NoC}}@\overline{\text{q}}$ refers to the average number of clicks needed to reach an IoU of q\% across all objects. Following previous works~\cite{Kontogianni2023InterObject3D,Yue2024AGILE3D}, the number of clicks per object is limited to a maximum of 10. For the \textit{4D setup}, we compute the average clicks across all superimposed \lidar{} scans to ensure comparability with the 3D setup. To compare with the recent work ClickFormer~\cite{Han2024ClickFormer}, we also adopt the mIoU@$\overline{\text{k}}$ metric. This metric is similar to IoU@k but it averages within each class before averaging across all classes. Also, in all experiments, during evaluation, we utilize the clicking strategy the model was trained with which is the most favorable condition for all methods.

\parag{In-distribution Evaluation.}
We first test our approach by training and evaluating on the same dataset.
As shown in Tab.~\ref{table:within_domain}, \ours{} achieves significant improvements over the current state-of-the-art, \agilethreed{}, in all metrics across all setups. Our \ours{}, which processes scenes holistically in 4D (see Tab.~\ref{table:within_domain}, \textit{last row}), surpasses all methods that operate on isolated 3D scans. Notably, with only \textbf{1 click} per object, it attains an \textbf{87.7\% $\overline{\text{IoU}}$}, outperforming \agilethreed{}'s \textbf{81.5\%}. Moreover, \ours{} reaches an \textbf{90\% $\overline{\text{IoU}}$}, a crucial threshold for annotation quality, with just \textbf{2.28 clicks}.

We also evaluate our 3D multi-object and 4D  models (introduced in Tab.~\ref{table:within_domain}) in the LPS and 4D-LPS tasks on \semantickitti{} and compare it with the state-of-the-art non-interactive method~\cite{Yilmaz2024Mask4Former},
which does not employ any human interaction. As seen in~Tab.~\ref{table:3d_4d_panoptic}, our interactive models significantly outperform this method, even with \textbf{1 click} per object, and continue to improve with refinement clicks.

\begin{table}[t]
\renewcommand{\arraystretch}{1.2}
\center
\tabcolsep=0.06cm
\footnotesize{
\caption{\small\textbf{Zero-shot Interactive \lidar{} Segmentation.}}
\label{table:zero_shot}
\begin{tabular}{cl c cccccc c ccc}
\toprule
\multicolumn{13}{c}{\textbf{Training}: \semantickitti{} $\rightarrow$ \textbf{Evaluation}: \nuscenes{}} \\
\midrule
\multicolumn{2}{c}{} & & \multicolumn{6}{c}{$\overline{\text{IoU}}\text{@}\overline{\text{k}}\uparrow$} & & \multicolumn{3}{c}{$\overline{\text{NoC}}\text{@}\overline{\text{q}}\downarrow$} \\
\cmidrule(lr){4-9} \cmidrule(lr){11-13}
& \textbf{Method} & & @1 & @2 & @3 & @4 & @5 & @10 & & @50 & @65 & @80\\
\midrule
\multirow{2}{*}{\rotatebox{90}{Single}} & \agilethreed~\cite{Yue2024AGILE3D} & & 32.4 & 40.8 & 47.1 & 52.2 & 56.4 & 68.4 & & 4.22 & 5.10 & 6.10 \\
& \cellcolor{gray!10}\textbf{\ours} & \cellcolor{gray!10} & \cellcolor{gray!10}45.5 & \cellcolor{gray!10}52.1 & \cellcolor{gray!10}57.2 & \cellcolor{gray!10}61.2 & \cellcolor{gray!10}64.6 & \cellcolor{gray!10}74.3 & \cellcolor{gray!10} & \cellcolor{gray!10}3.61 & \cellcolor{gray!10}4.40 & \cellcolor{gray!10}5.30 \\
\midrule
\multirow{2}{*}{\rotatebox{90}{Multi}} & \agilethreed~\cite{Yue2024AGILE3D} & & 32.5 & 37.4 & 42.4 & 47.8 & 52.7 & 66.0 & & 3.68 & 4.27 & 5.13 \\
& \cellcolor{gray!10}\textbf{\ours} & \cellcolor{gray!10} & \cellcolor{gray!10}44.2 & \cellcolor{gray!10}56.4 & \cellcolor{gray!10}63.3 & \cellcolor{gray!10}67.4 & \cellcolor{gray!10}70.2 & \cellcolor{gray!10}76.1 & \cellcolor{gray!10} & \cellcolor{gray!10}3.13 & \cellcolor{gray!10}3.76 & \cellcolor{gray!10}4.71 \\
\midrule
\multirow{2}{*}{\rotatebox{90}{4D}} & \agilethreed~\cite{Yue2024AGILE3D} & & 37.6 & 42.6 & 44.0 & 44.1 & 44.2 & 44.6 & & 3.96 & 4.64 & 5.66 \\
& \cellcolor{gray!10}\textbf{\ours} & \cellcolor{gray!10} & \cellcolor{gray!10}\textbf{54.7} & \cellcolor{gray!10}\textbf{58.1} & \cellcolor{gray!10}\textbf{63.7} & \cellcolor{gray!10}\textbf{69.5} & \cellcolor{gray!10}\textbf{73.2} & \cellcolor{gray!10}\textbf{79.6} & \cellcolor{gray!10} & \cellcolor{gray!10}\textbf{1.98} & \cellcolor{gray!10}\textbf{2.51} & \cellcolor{gray!10}\textbf{3.34} \\
\bottomrule
\end{tabular}
}
\end{table}

\begin{table}[t]
\renewcommand{\arraystretch}{1.2}
\center
\tabcolsep=0.03cm
\footnotesize{
\caption{\small\textbf{Comparison with previous works on 3D interactive single-object segmentation.} \textit{$\dag$: Zero-Shot Evaluation}.}
\label{table:clickformer}
\begin{tabular}{p{2.5cm} c cc c cc c cc c cc}
\toprule
\multicolumn{2}{c}{} & \multicolumn{2}{c}{mIoU@$\overline{\text{1}}\uparrow$} & & \multicolumn{2}{c}{mIoU@$\overline{\text{3}}\uparrow$} & & \multicolumn{2}{c}{mIoU@$\overline{\text{5}}\uparrow$} & & \multicolumn{2}{c}{mIoU@$\overline{\text{10}}\uparrow$}\\
\cmidrule(lr){3-4}\cmidrule(lr){6-7}\cmidrule(lr){9-10}\cmidrule(lr){12-13}
\textbf{Method} & & \textit{things} & \textit{stuff} & & \textit{things} & \textit{stuff} & & \textit{things} & \textit{stuff} & & \textit{things} & \textit{stuff} \\ 
\midrule
\multicolumn{12}{c}{{{{\textbf{Training}: \nuscenes{} $\rightarrow$ \textbf{Evaluation}: \nuscenes{}}}}}   \\
\arrayrulecolor{black!30}\midrule\arrayrulecolor{black}

CRSNet~\cite{Sun2023ACI} & & 31.2 & 7.1 & & 43.9 & 17.0 & & 46.4 & 22.7 & & 50.1 & 31.8 \\
InterObject3D~\cite{Kontogianni2023InterObject3D} & & 29.0 & 10.8 & & 44.6 & 25.7 & & 50.9 & 31.0 & & 55.6 & 38.6 \\
ClickFormer~\cite{Han2024ClickFormer} & & \textbf{35.2} & \textbf{48.6} & & 50.2 & 62.2 & & 56.9 & 64.3 & & 60.6 & 65.4 \\
$\dag$\cellcolor{gray!10} \agilethreed~\cite{Yue2024AGILE3D} & \cellcolor{gray!10} & \cellcolor{gray!10}28.7 & \cellcolor{gray!10}28.1 & \cellcolor{gray!10} & \cellcolor{gray!10}34.8 & \cellcolor{gray!10}46.6 & \cellcolor{gray!10} & \cellcolor{gray!10}44.2 & \cellcolor{gray!10}55.4 & \cellcolor{gray!10} & \cellcolor{gray!10}57.9 & \cellcolor{gray!10}63.9 \\
$\dag$ \cellcolor{gray!10}\textbf{\ours}&\cellcolor{gray!10} & \cellcolor{gray!10}35.0 & \cellcolor{gray!10} 43.6 &\cellcolor{gray!10} & \cellcolor{gray!10}\textbf{54.9} &\cellcolor{gray!10} \textbf{63.9} & \cellcolor{gray!10}& \cellcolor{gray!10}\textbf{62.1} & \cellcolor{gray!10}\textbf{70.8} & \cellcolor{gray!10}&\cellcolor{gray!10}\textbf{69.5} & \cellcolor{gray!10}\textbf{77.2} \\
\midrule
\multicolumn{12}{c}{{\textbf{Training}: \semantickitti{} $\rightarrow$ \textbf{Evaluation}: \kitti{}}}  \\
\arrayrulecolor{black!30}\midrule\arrayrulecolor{black}

CRSNet~\cite{Sun2023ACI} & & 28.3 & 9.6 & & 40.0 & 17.3 && 41.1 & 23.3 & &46.9 & 29.8 \\
InterObject3D~\cite{Kontogianni2023InterObject3D} & & 34.0 & 12.3 & & 42.6 & 23.1 & & 45.8 & 29.6 & & 49.6 & 36.1 \\
\agilethreed~\cite{Yue2024AGILE3D} & & 36.3 & 27.6 & & 47.3 & 44.0 & & 53.5 & 50.2 & & 63.3 & 59.6 \\
ClickFormer~\cite{Han2024ClickFormer} & & 28.0 & \textbf{41.1} & & 50.4 & 52.3 && 54.4 & 55.4 && 59.3 & 58.4 \\
\textbf{\ours} & & \textbf{47.7} & 39.5 && \textbf{59.4} & \textbf{55.7} & & \textbf{64.1} & \textbf{59.7} && \textbf{70.0} & \textbf{65.1} \\
\bottomrule
\end{tabular}
}
\end{table}

\parag{Zero-shot Evaluation.}
We assess the generalization capability of our method by evaluating it on the \nuscenes{} dataset, which significantly differs from \semantickitti{} in terms of environments and the \lidar{} sensor.
As shown in Tab.~\ref{table:zero_shot}, \ours{}  substantially outperforms the baseline, achieving nearly \textbf{70\% $\overline{\text{IoU}}$} with just \textbf{4 clicks} per object and further increasing by an additional 10\% with more clicks.

\newcommand{\numberone}{\normalsize\ding{192}}
\newcommand{\numbertwo}{\normalsize\ding{193}}
\newcommand{\numberthree}{\normalsize\ding{194}}
\newcommand{\numberfour}{\normalsize\ding{195}}
\newcommand{\numberfive}{\normalsize\ding{196}}
\newcommand{\numbersix}{\normalsize\ding{197}}
\newcommand{\numberseven}{\normalsize\ding{198}}
\newcommand{\numbereight}{\normalsize\ding{199}}
\newcommand{\numbernine}{\normalsize\ding{200}}
\newcommand{\numberten}{\normalsize\ding{201}}
\newcommand{\scissorright}{\normalsize\ding{36}}
\newcommand{\clubs}{\normalsize\ding{168}}
\newcommand{\spades}{\normalsize\ding{171}}
\begin{table}[t]
\renewcommand{\arraystretch}{1.2}
\center
\tabcolsep=0.06cm
\footnotesize{
\caption{\small\textbf{Click Simulations and Architectural Enhancements.}
\textit{BD: Boundary Dependent}-Eq.~\eqref{eq:former_clicking}. \textit{SI: Scale Invariant}-Eq.~\eqref{eq: si clicking}}.
\label{table:click_simulation_ablation}
\begin{tabular}{c ccc c ccc c cc}
\toprule
& \multicolumn{3}{c}{Clicking} & & \multicolumn{3}{c}{$\overline{\text{IoU}}\text{@}\overline{\text{k}}\uparrow$} & & \multicolumn{2}{c}{$\overline{\text{NoC}}\text{@}\overline{\text{q}}\downarrow$} \\
\cmidrule{2-4} \cmidrule{6-8}\cmidrule{10-11}
& \makecell{\hspace{0.5cm}SI} & \makecell{\hspace{0.5cm}Initial} & \makecell{Refinement} & & @1 & @5 & @10 & & @80 & @90 \\
\cmidrule{2-11}
\numberone & \hspace{0.5cm}\xmark & \hspace{0.5cm}BD & BD & & 70.1 & 83.3 & 86.6 & & 2.82 & 3.98\\
\numbertwo  & \hspace{0.5cm}\xmark & \hspace{0.5cm}Random & Random & & 72.3& 77.9 & 84.0  & & 3.17 & 4.45 \\
\arrayrulecolor{black!10}\cmidrule{2-11}\arrayrulecolor{black}
\numberthree & \hspace{0.5cm}\cmark & \hspace{0.5cm}BD & BD & & 33.6  & 72.2  & 79.1 & & 4.58 & 5.67 \\
\numberfour & \hspace{0.5cm}\cmark & \hspace{0.5cm}Random & Random & & 70.4  & 86.9 & 89.1 & & 2.81 & 3.94 \\
\numberfive & \hspace{0.5cm}\cmark & \hspace{0.5cm}Centroid & Centroid & & 75.1  & 85.6 & 88.0 & & 2.88 & 4.10 \\
\numbersix & \hspace{0.5cm}\cmark & \hspace{0.5cm}Centroid & BD & & 75.1 & 86.1  & 88.7 & & 2.87 & 4.00 \\
\tikz[remember picture] \node[inner sep=0pt,anchor=base] (n7) {\numberseven}; & \hspace{0.5cm}\cmark & \hspace{0.5cm}Centroid & Random & & 75.3 & 87.6 & 89.8  & & 2.72 & 3.84 \\
\arrayrulecolor{black!10}\cmidrule{2-11}\arrayrulecolor{black}
\multicolumn{4}{l}{\underline{Architectural Enhancements}} & & & & & & & \\
\tikz[remember picture] \node[inner sep=0pt,anchor=base] (n8) {\numbereight}; & \multicolumn{3}{l}{Identity Encoding} & & 75.8 & 85.8 & 90.4 & & 2.67 & 3.78 \\
\tikz[remember picture] \node[inner sep=0pt,anchor=base] (n9) {\numbernine}; & \multicolumn{3}{l}{Localized Loss} & & 76.6 & 86.3 & 90.6 & & 2.60 & 3.70 \\
\tikz[remember picture] \node[inner sep=0pt,anchor=base] (n10) {\numberten}; & \multicolumn{3}{l}{Identity Encoding + Localized Loss} & & \textbf{77.4} & \textbf{89.1} & \textbf{91.2} & & \textbf{2.53} & \textbf{3.62} \\
\bottomrule
\end{tabular}
\begin{tikzpicture}[overlay, remember picture,>=Stealth]
\draw[->, arrowgray,rounded corners=2pt] (n7.west) -- ++(-0.3cm,0) |- (n8.west);
\draw[->, arrowgray,rounded corners=2pt] (n7.west) -- ++(-0.3cm,0) |- (n9.west);
\draw[->, arrowgray,rounded corners=2pt] (n7.west) -- ++(-0.3cm,0) |- (n10.west);
\end{tikzpicture}
}

\end{table}

We also compare our approach with CRSNet~\cite{Sun2023ACI} and ClickFormer~\cite{Han2024ClickFormer}, which are limited to 3D single-object setup. Due to the lack of public codebases, we could not adapt these methods to 3D multi-object or 4D setups and followed their original evaluation protocols. As shown in Tab.~\ref{table:clickformer}, \ours{} significantly outperforms these methods on the \nuscenes{} dataset, \underline{despite not being trained on it}, unlike the baselines.
We also evaluate our model on the KITTI-360~\cite{liao2022kitti360} dataset for a fair comparison. \ours{} consistently outperforms all baselines on KITTI-360, further demonstrating its robustness when applied to a new dataset. The results for competitors are taken from~\cite{Han2024ClickFormer}.

\parag{Ablation Studies.}
All ablations are conducted on \textit{\indomain{}} 3D multi-object interactive segmentation setup.

\subsubsection{\textit{Click Simulation Strategy}}\label{sec:click_simulation} In Tab.~\ref{table:click_simulation_ablation}, we demonstrate the effectiveness of our proposed click simulation strategy (see Sec.~\ref{sec:click_sim}) compared to different clicking strategies. 
As shown in Tab.~\ref{table:click_simulation_ablation}-\numberseven{}, our strategy outperforms all others, particularly \numberone{} and \numbertwo{}%
, which are used in previous interactive methods. With just \textbf{1 click}, our approach achieves a better \textbf{$\overline{\text{IoU}}$}~(\textbf{75.3\%} vs. \textbf{70.1\%}) maintaining strong performance at higher click counts (\textbf{89.8\% $\overline{\text{IoU}}@\overline{\text{10}}$}). These results highlight that existing strategies, which perform well in dense domains, do not transfer effectively to sparse \lidar{}.

We also assessed the robustness of interactive methods when click simulation strategies differ between training and testing.
We used \agilethreed{} clicking strategy \underline{only} during the testing of \ours{}. This resulted in an 87.0\% $\overline{\text{IoU}}@\overline{\text{10}}$, outperforming \agilethreed{}'s 86.6\%, even though \agilethreed{} was trained with the \underline{same} clicking strategy.
Moreover, we used our clicking strategy during the testing of \agilethreed{}, which demonstrated a bigger drop from 86.6\% to 81.2\% $\overline{\text{IoU}}@\overline{\text{10}}$ (5.4 points). %
All of these shows that \ours{} is more robust against different clicking strategies, and our improvements not only originate from the click simulation but also from our architectural design and training.

\subsubsection{Robustness to Click Randomness}
\label{sec:click_randomness}
Our method also incorporates randomness in click selection. To assess its sensitivity to this randomness, we repeated all experiments three times and calculated the standard deviation (std) of the results. We report the average results. The std was minimal (for $\overline{\text{IoU}}@\overline{\text{k}}$  $\approx 0.04$ and for $\overline{\text{NoC}}@\overline{\text{q}} \approx 0.004$), confirming the method's robustness to click selection. 

\begin{figure}[t]
  \begin{minipage}[t]{.45\linewidth}
    \captionof{figure}{\small\textbf{Number of Superimposed Scans Ablation.}}
    \vspace{-0.1cm}
    \definecolor{my_blue}{RGB}{38,134,245}
\begin{tikzpicture}
    \begin{axis}[
        width=5cm,
        height=5cm,
        xlabel={\scriptsize{Number of superimposed scans}},
        enlarge y limits = true,
        tick label style={font=\scriptsize},
        xmin=1, xmax=10,
        ymin=75, ymax=91,
        legend style={font=\scriptsize, at={(0.90,0.15)}, anchor=south east},
        legend cell align={left},
    ]
    \addplot[
        color=my_blue,
        mark=*,
        mark options={fill=my_blue},
        line width=0.3mm
    ] coordinates {
        (1, 77.4) (2,85.3) (4,88) (8,88.9) (10,89.4)
    };
    \addlegendentry{$\overline{\text{IoU}}@\overline{1}$};
    
    \node[font=\scriptsize, color=my_blue] at (80,25) {77.4};
    \node[font=\scriptsize, color=my_blue] at (100,115) {85.3};
    \node[font=\scriptsize, color=my_blue] at (300,142) {88.0};
    \node[font=\scriptsize, color=my_blue] at (680,152) {88.9};
    \node[font=\scriptsize, color=my_blue] at (835,155) {89.4};

    \end{axis}
\end{tikzpicture}
    \label{fig:ablation_overlap}
  \end{minipage}\hspace{0.30cm}
  \begin{minipage}[t]{.45\linewidth}

    \renewcommand{\arraystretch}{1.2}
    \setlength{\tabcolsep}{4pt}
    \centering
    \footnotesize{
    \captionof{table}{\small\textbf{User Study with Human Annotators.}
    }
    \label{table:user_study}
    \begin{tabular}{l c c}

    \toprule
    \multicolumn{3}{c}{\textbf{\semantickitti{}}}\\
    \midrule
    & $\overline{\text{IoU}}@\overline{\text{3}}\uparrow$ & $\overline{\text{t}}$\\
    \midrule
    Human &  $95.0\pm{0.4}$ & 5 min. \\
    Simulator &  94.1 & -- \\
    \midrule
    \multicolumn{3}{c}{\textbf{\nuscenes{}}}\\
    \midrule
    & $\overline{\text{IoU}}@\overline{\text{7}}\uparrow$ & $\overline{\text{t}}$\\
    \cmidrule{1-3}
     Human &  $90.8\pm{2.3}$& 6 min. \\
     Simulator &  91.5 & -- \\
    \bottomrule
    \end{tabular}}
  
  \end{minipage}
  \vspace{-0.5cm}
\end{figure}

\begin{figure}[ht!]
\centering
\includegraphics[width=1\linewidth]{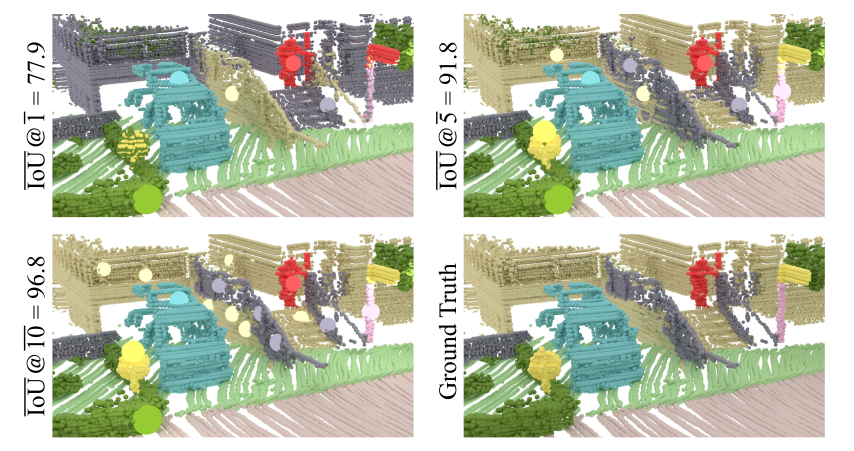}
\vspace{-0.5cm}
\caption{\small{Example results of \ours{} on SemanticKITTI.}}
\label{fig:qualitatie_resusts}
\end{figure}

\subsubsection{\textit{Architectural Enhancements}}
Our additional structural modifications to the model led to further key improvements (\numberseven-\numberten). As shown in Table~\ref{table:click_simulation_ablation}, both the identity encoding (\numbereight), which injects learned encodings into object queries, and localizing the loss (\numbernine) enhance the model performance. Each component allows us to reach the 90\% threshold. When combined (\numberten), these enhancements result in a significant improvement, especially considering the already high baseline, where further gains are typically more challenging.

\subsubsection{{Number of Superimposed Scans in 4D}}
In Tables~\ref{table:within_domain},\ref{table:3d_4d_panoptic}, \ref{table:zero_shot}, all 4D results are presented with 4 superimposed scans to ensure comparability with \agilethreed{}, which faces memory limitations beyond 4 scans. Fig.~\ref{fig:ablation_overlap} shows that superimposing more scans increases the segmentation accuracy of \ours{} even more, especially at lower click counts.

\parag{User Study.} To assess the practical applicability of \ours{} for annotating \lidar{} data, we integrated \ours{} to a user interface~\cite{Yue2024AGILE3D} and conducted a user study with ten human annotators. They interactively segmented selected scenes from \semantickitti{} and \nuscenes{}, each consisting of 4 consecutive scans. The annotators had no prior annotating experience and were given no clicking instructions. Tab.~\ref{table:user_study} summarizes the results. The annotators achieve comparable results to the simulator. This proves \ours{} not only performs well with simulated clicks but also with real annotators.

\parag{Qualitative Results}
are shown in Fig.~\ref{fig:qualitatie_resusts}. \ours{} effectively segments and refines cluttered scenes.

\newcommand{\setsectionspacing}{
  \vspace{-3pt}
}
\setsectionspacing
\section{CONCLUSION}

We have introduced interactive 4D segmentation, a new paradigm where a user segments multiple objects on multiple scans simultaneously and \ours{}, the first interactive 4D segmentation method following this paradigm. \ours{} is significantly more efficient than the previous approaches, limited to a single object and single scan. Along with our new click simulation strategy suited for sparse \lidar{} scans, it also shows outstanding results in accuracy and reaches state-of-the-art performance by a large margin. We hope \ours{} will reduce the annotation effort required for future \lidar{} datasets.

{\parag{Acknowledgments.} We thank Yuanwen Yue, Daan de Geus, and Alexander Hermans for their helpful feedback and discussions. We also thank all our annotators who participated in the user study. Theodora Kontogianni is a postdoctoral research fellow at the ETH AI Center and her research is partially funded by the Hasler Stiftung Grant project (\texttt{23069}). Idil Esen Zulfikar’s research is funded by the BMBF project NeuroSys-D (\texttt{03ZU1106DA}). Kadir Yilmaz's research is funded by the Bosch-RWTH LHC project Context Understanding for Autonomous Systems. The computing resources for most of the experiments were granted by the Gauss Centre for Supercomputing e.V. through the John von Neumann Institute for Computing on the GCS Supercomputer JUWELS at Julich Supercomputing Centre.}

\bibliographystyle{plain}
\bibliography{abbrev,egbib}

\clearpage

\twocolumn[{
\renewcommand\twocolumn[1][]{#1}
\vspace{0.1cm}

\begin{center}
\textbf{\Large Interactive4D: Interactive 4D \lidar{} Segmentation\\[0.2cm]
\textit{Supplementary Material}}
\end{center}
\vspace{0.5cm}
}]

\textit{This supplementary document is structured as follows:}
\begin{itemize}
    \item § \ref{sec:implementation_details}: \nameref{sec:implementation_details}
    \begin{itemize}
        \item \ref{subsec: point cloud construction}: \nameref{subsec: point cloud construction}
        \item \ref{subsec: evaluation metric extension}: \nameref{subsec: evaluation metric extension}
        \item \ref{subsec: iterative training}: \nameref{subsec: iterative training}
        \item \ref{subsec: dataset details}: \nameref{subsec: dataset details}
        \item \ref{subsec: training details}: \nameref{subsec: training details}
        \item \ref{subsec: user study details}: \nameref{subsec: user study details}
    \end{itemize}
    \item § \ref{sec:limitation}: \nameref{sec:limitation}
    \begin{itemize}
        \item \ref{subsec: memory demand}: \nameref{subsec: memory demand}
        \item \ref{subsec: run time}: \nameref{subsec: run time}
        \item \ref{subsec: tracking failure}: \nameref{subsec: tracking failure}
    \end{itemize}
    \item § \ref{sec:additional_ablation}: \nameref{sec:additional_ablation}
        \begin{itemize}
        \item \ref{subsec: dbscan}: \nameref{subsec: dbscan}
        \item \ref{subsec: click budget}: \nameref{subsec: click budget}
        \item \ref{subsec: voxel size}: \nameref{subsec: voxel size}
        \item \ref{subsec: time ablation}: \nameref{subsec: time ablation}
    \end{itemize}
    \item § \ref{subsec: additional results}: \nameref{subsec: additional results}
        \begin{itemize}
        \item \ref{subsec: class wise results}: \nameref{subsec: class wise results}
        \item \ref{subsec: additional qualitative results}: \nameref{subsec: additional qualitative results}
    \end{itemize}
\end{itemize}

\section{Implementation Details.} \label{sec:implementation_details}
\subsection{Spatio-Temporal Point Cloud Construction} \label{subsec: point cloud construction}
Following the methodology of prior work \cite{Yilmaz2024Mask4Former}, we leverage poses of the ego vehicle to transform all the \lidar{} scans into a global coordinate frame, creating a consistent representation. Then, we superimpose a short temporal window of $T$ consecutive \lidar{} scans into a single spatio-temporal point cloud $\mathcal{P}_t$~$\in$~$\mathbb{R}^{M\times3}$. 

To formalize, let $\mathbf{P}_i \in \mathbb{R}^{N_i \times 3}$ represent the point cloud from the $i$-th \lidar{} scan, where $N_i$ is the number of points in the point cloud, and each point has 3 coordinates $(x, y, z)$. To transform each point cloud $\mathbf{P}_i$ from the local coordinate frame of the ego vehicle to the global coordinate frame, we apply a transformation matrix $\mathbf{T}_i$, which consists of a rotation matrix $\mathbf{R}_i \in \mathbb{R}^{3 \times 3}$ and a translation vector $\mathbf{t}_i \in \mathbb{R}^{3}$. The transformation is given by:
\begin{equation}
\mathbf{P}_i^{\text{global}} = \mathbf{P}_i \mathbf{R}_i^\top + \mathbf{t}_i
\end{equation}
where $\mathbf{P}_i \mathbf{R}_i^\top$ applies the rotation to all points in $\mathbf{P}_i$, and $\mathbf{t}_i$ translates the points into the global coordinate frame  as in \cite{Yilmaz2024Mask4Former}.
After transforming all point clouds $\mathbf{P}_1^{\text{global}}, \mathbf{P}_2^{\text{global}}, \dots, \mathbf{P}_{\tau}^{\text{global}}$ from a temporal window of $\tau$ consecutive scans, we aggregate them into a unified spatio-temporal point cloud $\mathbf{P}^{\text{ST}} \in \mathbb{R}^{ \left(\sum_{i=1}^{\tau} N_i\right) \times 3}$, which is formed by stacking all the transformed point clouds:
\begin{equation}
\mathbf{P}^{\text{ST}} = \bigcup_{i=1}^{\tau} \mathbf{P}_i^{\text{global}}
\end{equation}

\begin{table*}[h]
\renewcommand{\arraystretch}{1.2}
\setlength{\tabcolsep}{7pt}
\footnotesize{
\centering
\caption{\textbf{\ours{} and \agilethreed{} per class in-distribution interactive \lidar{} segmentation results.} The results are reported for $\overline{\text{IoU}}@\overline{\text{10}}$.}
\label{table:semantickitti_objects}
\resizebox{\textwidth}{!}{
\begin{tabular}{cc|cccccccccccccccccccccc}
\toprule
\multicolumn{21}{c}{\textbf{Training}: \semantickitti{}  $\rightarrow$ \textbf{Evaluation}: \semantickitti{}} \\
\midrule
& Method & \multicolumn{19}{c}{Categories} \\

\midrule
& & \multicolumn{8}{c}{\textit{Things}} & \multicolumn{11}{c}{\cellcolor{gray!10}\textit{Stuff}} \\
\cmidrule(r){3-10} \cmidrule(l){11-21}

& & {\rotatebox[origin=c]{90}{Car}} & {\rotatebox[origin=c]{90}{Bicycle}} & {\rotatebox[origin=c]{90}{Motorcycle}} & {\rotatebox[origin=c]{90}{Truck}} & {\rotatebox[origin=c]{90}{Other-vehicle}}  & {\rotatebox[origin=c]{90}{Person}} & {\rotatebox[origin=c]{90}{Bicyclist}} & {\rotatebox[origin=c]{90}{Motorcyclist}} & {\cellcolor{gray!10} \rotatebox[origin=c]{90}{Road}} & { \cellcolor{gray!10} \rotatebox[origin=c]{90}{Parking}} & { \cellcolor{gray!10}\rotatebox[origin=c]{90}{Sidewalk}} & {\cellcolor{gray!10}\rotatebox[origin=c]{90}{Other-ground}}  & {\cellcolor{gray!10}\rotatebox[origin=c]{90}{Building}} & {\cellcolor{gray!10}\rotatebox[origin=c]{90}{Fence}} & {\cellcolor{gray!10}\rotatebox[origin=c]{90}{Vegetation}} & {\cellcolor{gray!10}\rotatebox[origin=c]{90}{Trunk}} & {\cellcolor{gray!10}\rotatebox[origin=c]{90}{Terrain}}  & {\cellcolor{gray!10}\rotatebox[origin=c]{90}{Pole}} & {\cellcolor{gray!10}\rotatebox[origin=c]{90}{Traffic-sign}}\\

\midrule

\multirow{2}{*}{\rotatebox{90}{Single}} & \agilethreed~\cite{Yue2024AGILE3D} & 96.1 & 73.5 & 85.8 & 96.0 & 85.2 & 78.9 & 93.6 & 66.1 & 89.4 & 47.6 & 69.7 & 55.8 & 84.8 & 47.2 & 84.6 & 49.1 & 64.0 & 54.9 & 69.4 \\
 & \textbf{\ours} & \textbf{97.8} & 80.2 & 91.9 & 96.4 & 92.0 & 87.4 & \textbf{95.9} & 85.4 & 92.2 & 62.5 & 77.8 & \textbf{72.5} & 90.8 & 59.9 & 85.1 & 66.5 & 69.6 & 69.6 & 83.2 \\
\midrule

\multirow{2}{*}{\rotatebox{90}{Multi}} & \agilethreed~\cite{Yue2024AGILE3D} & 94.4 & 66.3 & 85.7 & 97.0 & 87.4 & 77.3 & 91.3 & 83.4 & 94.3 & 74.2 & 84.0 & 63.1 & 93.4 & 58.2 & 89.4 & 75.2 & 73.3 & 72.0 & 85.0 \\
& \textbf{\ours} & 95.5 & 80.4 & 92.6 & 97.7 & 92.6 & 86.2 & 92.5 & \textbf{92.9} & \textbf{95.7} & \textbf{81.7} & \textbf{89.2} & 70.3 & 94.5 & 76.4 & 92.9 & 83.7 & \textbf{82.6} & 86.1 & 90.9 \\
\midrule

\multirow{2}{*}{\rotatebox{90}{4D}} & \agilethreed~\cite{Yue2024AGILE3D} & 92.1 & 60.8 & 82.1 & 96.4 & 85.4 & 73.7 & 89.0 & 71.5 & 94.5 & 74.4 & 86.0 & 52.1 & 95.8 & 63.7 & 92.7 & 77.1 & 77.3 & 72.4 & 66.3 \\
& \textbf{\ours} & 96.3 & \textbf{90.0} & \textbf{95.0} & \textbf{98.3} & \textbf{94.9} & \textbf{93.6} & 95.0 & 92.2 & 95.2 & 75.7 & 87.5 & 63.7 & \textbf{96.4} & \textbf{79.9} & \textbf{93.8} & \textbf{87.1} & 81.8 & \textbf{89.8} & \textbf{95.0} \\
\bottomrule
\end{tabular}
}
}
\end{table*}

\subsection{Details on Evaluation Metrics} \label{subsec: evaluation metric extension}
In this section, we explain the evaluation metrics $\overline{\text{IoU}}@\overline{\text{k}}$ and $\overline{\text{NoC}}@\overline{\text{q}}$ in detail, particularly how they are adapted for the 4D setup to ensure a fair comparison with the 3D setup.

\textbf{Evaluation Click Budget:} To ensure consistency between the 3D and 4D setups, we maintain the same total number of clicks across the entire dataset. In the 3D setup, the click budget for each scan $i$ is given by $B \times O_i$, where $O_i$ is the number of objects in scan $i$, and $B$ is the click budget per object. However, in the 4D setup, we no longer operate within the confines of a single scan but instead use a temporal window of $\tau$ scans. Therefore, for a given temporal window $t$, the click budget $B_t$ is the sum of the budgets for each scan $i$ within the window:
$$ B_t = \sum_{i=t}^{t+\tau} B \times O_i $$

\textbf{$\overline{\text{IoU}}@\overline{\text{k}}$:} In both 3D and 4D setups, the IoU metric is computed by averaging the object IoUs across the entire validation set, where each object in each \lidar{} scan is treated as an individual entity. In the 3D setup, this is straightforward as predictions are made directly in single \lidar{} scans. However, in the 4D setup, predictions are made on superimposed point clouds. To ensure comparability with the 3D setup, we split the combined point cloud back into individual \lidar{} scans and calculate the object IoU for each scan separately.
Furthermore, a fair $\overline{\text{IoU}}@\overline{\text{k}}$ comparison between the 3D and 4D setups requires the same amount of clicks (k) per object for the entire annotation process. To ensure this, in the 4D setup, we report the $\overline{\text{IoU}}@\overline{\text{k}}$ for an average of k clicks given per object per \lidar{} scan. As a result, in the superimposed point clouds $k \times \tau$ clicks are given per object as the user segments $\tau$ \lidar{} scans at the same time. This ensures that in both 3D and 4D setups, at each reported click count k, the same number of clicks are given over the entire sequence, allowing for a direct comparison of results.

\textbf{$\overline{\text{NoC}}@\overline{\text{q}}$:}
First of all, it is important to highlight that, to be consistent with the prior works~\cite{Kontogianni2023InterObject3D, Yue2024AGILE3D, Sun2023ACI, mahadevan2023itis, li2018interactive, mahadevan2023itis, kontogianni2020continuous, chen2022focalclick, sofiiuk2022reviving}, when an object does not achieve the q\% IoU threshold, even when maximum click budget B is given, we use B for $\overline{\text{NoC}}@\overline{\text{q}}$ calculation. This avoids excessive penalization (i.e., infinite) for objects that cannot reach the threshold, however, is misleading as these objects definitely require more than B clicks to achieve the desired threshold.

In the 3D setup, to calculate $\overline{\text{NoC}}@\overline{\text{q}}$, each object in each \lidar{} scan is attached with a click counter.
This counter starts from 0 and increases every time the object is given a click.
When the desired threshold is achieved, the current counter state is used for $\overline{\text{NoC}}@\overline{\text{q}}$ calculation.
To maintain fairness, following the rationale used in $\overline{\text{IoU}}@\overline{\text{k}}$, in the 4D setup, a click counter is attached to each object tracklet throughout the temporal window. Every time a click is given to an object tracklet in the superimposed point cloud, the counter increases. Once an object reaches the threshold in any one of the \lidar{} scans, the $\overline{\text{NoC}}@\overline{\text{q}}$ for this specific scan for that object is finalized as the current counter state, and the counter is reset to avoid double-counting those clicks for the next scans where the threshold will be reached. This prevents unfairly inflating click counts compared to the 3D setup and makes the setups comparable.
This boils down to averaging the total clicks over the \lidar{} scans and allows us to fairly assess the efficiency of the 4D setup. For example, if an object is successfully segmented across all scans with a single click, the $\overline{\text{NoC}}@\overline{\text{q}}$ for that object will be 1 in the first scan, and 0 in the subsequent scans, as no additional clicks were required.
Conversely, if an object fails to meet the threshold in a given scan, the maximum budget $B$ will be counted as the number of scans needed to reach this threshold. This method provides a fair evaluation of the 4D setup’s ability to minimize the number of clicks over time while acknowledging the computational advantage of segmenting multiple scans simultaneously.

\subsection{Iterative Training} \label{subsec: iterative training}
We adopt a multi-object iterative training approach, inspired by \agilethreed~\cite{Yue2024AGILE3D}, to enhance model performance by simulating user interaction. Specifically, multiple forward passes are performed for each batch to ensure that meaningful loss is maintained before backpropagation. This process mimics how a real user would iteratively provide feedback through clicks.

In each batch, we randomly select the number of iterations, $n$, from the range $[1, N_{max}]$, where $N_{max}$ is the maximal click budget per object. This ensures the model learns to perform well even when the click budget is not fully exhausted. Initially, clicks ($S^0$) are placed at the center of each target object, enabling the model to make an initial prediction. This prediction is then compared to the ground truth to identify error clusters. During subsequent iterations ($k$), to progressively guide the model toward more accurate predictions, the model’s performance is improved by sampling clicks ($C^k$), placing one click for each of the $|C_i|$ largest error regions. Sampling $N_i$ clicks together accelerates training, by reducing the number of forward passes required to reach the desired budget, thus keeping the computational complexity manageable.

The model remains frozen during click sampling from iterations 1 to $N_{iter}-1$, with backpropagation occurring only after the final iteration, reducing the overall computational cost.. This approach contrasts with full iterative training, which updates the model after each click. While full iterative training is used during testing, it is computationally expensive, as it requires frequent updates.

\begin{table*}[h]
\renewcommand{\arraystretch}{1.2}
\setlength{\tabcolsep}{4pt}
\centering
\caption{\textbf{\ours{} and \agilethreed{} per class zero-shot interactive \lidar{} segmentation results.} The results are reported for $\overline{\text{IoU}}@\overline{\text{10}}$.}
\label{table:nuscenes_objects}
\resizebox{\textwidth}{!}{
\begin{tabular}{cc|cccccccccccccccccccccccccccc}
\toprule
\multicolumn{30}{c}{\textbf{Training}: \semantickitti{}  $\rightarrow$ \textbf{Evaluation}: \nuscenes{}} \\
\midrule
& Method & \multicolumn{28}{c}{Categories} \\

\midrule
& & \multicolumn{20}{c}{\textit{Things}} & \multicolumn{8}{c}{\cellcolor{gray!10}\textit{Stuff}} \\
\cmidrule(r){3-22} \cmidrule(l){23-30}

 & & {\rotatebox[origin=c]{90}{animal}} & {\rotatebox[origin=c]{90}{adult}} &{\rotatebox[origin=c]{90}{child}} & {\rotatebox[origin=c]{90}{construction-worker}} & {\rotatebox[origin=c]{90}{police-officer}} & {\rotatebox[origin=c]{90}{stroller}} & {\rotatebox[origin=c]{90}{barrier}} & {\rotatebox[origin=c]{90}{debris}} & {\rotatebox[origin=c]{90}{pushable-pullable}} & {\rotatebox[origin=c]{90}{traffic-cone}} & {\rotatebox[origin=c]{90}{bicycle-rack}} & {\rotatebox[origin=c]{90}{bicycle}} & {\rotatebox[origin=c]{90}{bendy}} & {\rotatebox[origin=c]{90}{rigid}} & {\rotatebox[origin=c]{90}{car}} & {\rotatebox[origin=c]{90}{construction}} &  {\rotatebox[origin=c]{90}{police}} & {\rotatebox[origin=c]{90}{motorcycle}} & {\rotatebox[origin=c]{90}{trailer}} & {\rotatebox[origin=c]{90}{truck}} & {\cellcolor{gray!10}\rotatebox[origin=c]{90}{driveable-surface}} & {\cellcolor{gray!10}\rotatebox[origin=c]{90}{flat-other}} & {\cellcolor{gray!10}\rotatebox[origin=c]{90}{sidewalk}} & {\cellcolor{gray!10}\rotatebox[origin=c]{90}{terrain}} & {\cellcolor{gray!10}\rotatebox[origin=c]{90}{manmade}} & {\cellcolor{gray!10}\rotatebox[origin=c]{90}{static-other}} & {\cellcolor{gray!10}\rotatebox[origin=c]{90}{vegetation}} & {\cellcolor{gray!10}\rotatebox[origin=c]{90}{vehicle-ego}} \\
\midrule
\multirow{2}{*}{\rotatebox{90}{Single}} & \agilethreed~\cite{Yue2024AGILE3D}  & 54.7 & 71.8 & 55.9 & 72.1  & 95.5 & 77.8 & 57.3 & 83.7 & 68.3 & 62.9 & 59.6 & 77.8 & 78.1 & 82.7 & 87.6 & 75.0 & 92.8 & 82.8 & 68.7 & 68.7 & 77.1 & 24.2 & 27.0 & 24.6 & 20.0 & 50.2 & 56.8 & 98.9 \\ 
& \textbf{\ours}                                                              & 65.8 & 79.9 & 70.9 & \textbf{81.7}  & \textbf{95.3} & \textbf{84.7} & 67.1 & \textbf{91.7} & 73.1 & 73.4 & 69.4 & \textbf{84.0} & \textbf{85.8} & \textbf{91.9} & \textbf{94.3} & \textbf{88.7} & \textbf{94.7} & \textbf{91.6} &\textbf{ 82.3} & 82.3 & 83.2 & 39.8 & 42.1 & 41.1 & 20.8 & 64.4 & 21.4 & 94.2\\ 
\midrule
\multirow{2}{*}{\rotatebox{90}{Multi}} & \agilethreed~\cite{Yue2024AGILE3D}  & 19.3 & 50.7 & 45.6 & 51.4 & 74.6 & 63.0 & 57.8 & 62.4 & 38.3 & 49.3 & 47.2 & 44.5 & 69.1 & 74.0 & 78.1 & 66.4 & 86.8 & 56.7 & 74.6 & 74.6 & 86.4 & 41.6 & 56.0 & 52.6 & 60.7 & 51.0 & 72.6 & 90.6 \\ 
& \textbf{\ours}                                                              & 39.3 & 67.9 & 61.1 & 60.5 & 75.3 & 69.6 & 67.7 & 74.5 & 73.2 & 61.7 & 63.4 & 59.2 & 81.7 & 81.9& 82.8 & 75.0 & 86.8 & 75.6 & 78.9 & 78.9 & \textbf{94.2} & \textbf{66.3} & \textbf{72.7} & \textbf{70.2} & 75.3 & 61.1 & \textbf{78.5} & \textbf{99.0}\\ 
\midrule
\multirow{2}{*}{\rotatebox{90}{4D}} & \agilethreed~\cite{Yue2024AGILE3D} & 11.1 & 31.7 & 18.6 & 26.4  & 67.0 & 26.7 & 18.1 & 39.7 & 22.4 & 24.2 & 38.6 & 32.7 & 59.1 & 57.9 & 52.4 & 47.8 & 62.8 & 36.2 & 62.8 & 55.3 & 80.3 & 17.2 & 38.0 & 37.2 & 73.4 & 38.3 & 74.0 & 87.0 \\
& \textbf{\ours}                                                         & \textbf{83.9} & \textbf{81.7} & \textbf{81.9} & 76.3  & 91.6 & 82.9 & \textbf{72.3} & 88.0 & \textbf{80.8} & \textbf{83.9} & \textbf{71.5} & 78.7 & 76.0 & 82.7 & 84.2 & 81.1 & 80.0 & 82.2 & 80.0 & \textbf{82.5} & 87.9 & 50.0 & 52.8 & 56.3 & \textbf{78.9} & \textbf{71.3} & 77.6 & 94.3 \\
\bottomrule
\end{tabular}
}
\end{table*}

\subsection{Datasets Details} \label{subsec: dataset details}
\textbf{SemanticKITTI} is derived from the KITTI odometry dataset~\cite{Geiger2012AreWR}. The dataset consists of over $43,000$ \lidar{} scans recorded with a Velodyne-$64$ laser scanner capturing various urban driving scenarios at 10 Hz. It is split into training, validation, and test splits. Each point in the \lidar{} point clouds is densely annotated with one of $C$$=$$19$ semantic labels, \eg \emph{car}, \emph{road}, \emph{cyclist}, as well as a unique instance ID that is consistent over time. The dataset includes precise pose estimates of the ego vehicle for every time step, which is critical for the 4D interactive segmentation task.

\textbf{nuScenes} is a comprehensive dataset that includes sensor data from 1,000 diverse driving records, each lasting approximately 20 seconds. It is recorded in urban environments across Boston and Singapore. The dataset features data from various sensors including a 32-channel \lidar{} sensor, all synchronized and captured at a frequency of 20 Hz. Each point in the \lidar{} point clouds is densely annotated with one of $C$$=$$32$ semantic labels, \eg \emph{construction vehicle}, \emph{sidewalk}, \emph{motorcycle}, as well as a unique instance ID that is consistent over time. However, the annotations are provided only at 2 Hz, meaning that only 1 in 10 scans is annotated. The dataset offers precise pose estimates of the ego vehicle at every time step. In total  58,501 \lidar{} scans out of which 5850 are annotated.
We evaluate our models on the validation split that contains 150 sequences.

It is worth noting the key differences between these datasets. \semantickitti{} was collected in Karlsruhe, Germany, focusing on suburban areas with limited environmental diversity, whereas \nuscenes{} covers more varied and dynamic urban settings in Boston and Singapore, including different road types, weather conditions, and traffic densities.
In addition, the datasets differ in \lidar{} sensor specifications, such as sensor frequency and the number of laser beams, posing additional challenges for model generalization.

\textbf{KITTI-360} dataset is used to compare with other recent works~\cite{Sun2023ACI, Han2024ClickFormer} that do not have publicly available codebases. However, the dataset lacks per-point labels for each \lidar{} scan, providing labels only for down-sampled superimposed point clouds. To address this limitation and also be able to evaluate our model, we applied a nearest-neighbor algorithm~\cite{fix1951discriminatory} to propagate labels to individual points. Concretely, we use publicly available scripts (Sanchez, \href{https://github.com/JulesSanchez/recoverKITTI360label}{2021}). While evaluating our model, we use the entire validation split, both static and dynamic points provided by the dataset.

\subsection{Training Details} 
\label{subsec: training details}
Each spatiotemporal point cloud is formed by superimposing 4 consecutive \lidar{} scans that are voxelized with a voxel size of $10$~cm. For consistency, we set the click budget to 10 clicks per object both during training and evaluation. We use a Minkowski Res16UNet34C~\cite{choy20194d} as the sparse feature backbone. Our model is trained for 30 epochs with an effective batch size of 16 using the AdamW optimizer~\cite{Loshchilov2017DecoupledWD} and the one-cycle learning rate scheduler~\cite{Smith2017SuperconvergenceVF} with a maximum learning rate of $2\cdot10^{-4}$. We perform random rotation, translation, and scaling data augmentations. The training experiments are conducted on 16 NVIDIA A100 (40GB) GPUs, and the evaluation experiments on a single NVIDIA A40 (48GB) GPU. 

\subsection{User Study Details} 
\label{subsec: user study details} 
To assess the practical usability of our annotation approach, we conducted a user study involving 10 participants with no prior experience with the annotation process. We allowed users to follow their preferences for clicking on error regions, rather than explicitly instructing them to select the regions with the maximum error. The participants were provided with a brief written explanation of the 4D scene setup, including the dynamic nature of the silhouettes and their task of annotating objects in the given scenes. Each participant was given a budget of 10 clicks per object and instructed to continue annotating until they were satisfied with the results. Upon completion of each scene, the average intersection-over-union~(IoU) score was recorded, along with the number of clicks. The results reported in Tab.~\ref{table:user_study} were averaged across users to report the mean IoU per scene, as well as the average number of clicks used per object per participant.

The user study was conducted on four scenes where two of the scenes are selected from \semantickitti{} and the other two from \nuscenes{}. Also, each scene consists of four consecutive scans. 
To keep the study manageable in terms of time and GPU memory requirements, the participants were asked to annotate only 10 objects per scene. These objects included both dynamic and static categories, covering both \textit{stuff} and \textit{things}.

\begin{figure}[ht!]
\centering
\caption{Annotation tool user interface.}
\includegraphics[width=1\linewidth]{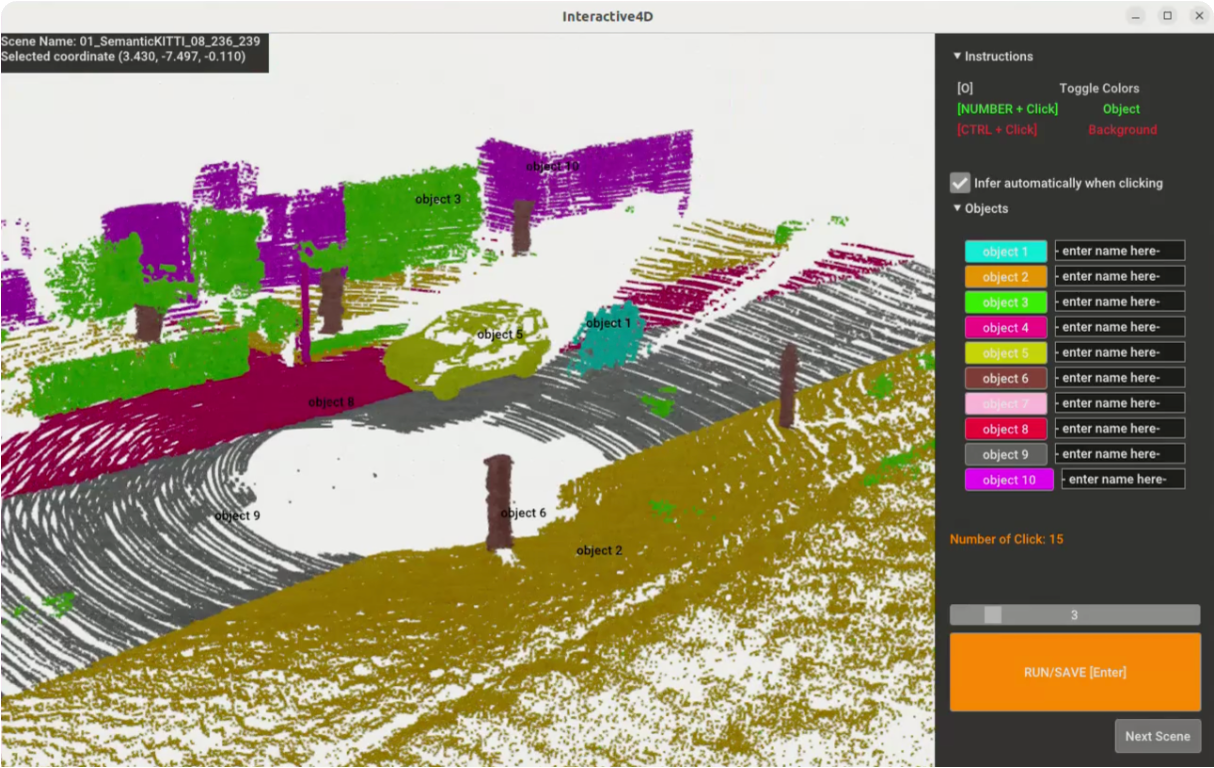}
\label{fig:annotation_tool}
\end{figure}

The annotation tool used for the study is shown in Fig.~\ref{fig:annotation_tool}. It is based on the \agilethreed~\cite{Yue2024AGILE3D} interface with several modifications to accommodate outdoor environments.
The tool enables users to modify the size of the points in the point cloud, allowing them to adjust point sizes based on their preferences, depending on the viewpoint, level of detail, or area of interest. Additionally, instead of requiring precise clicks on individual points, the tool provides a mechanism to infer broader regions based on user input, streamlining the interaction process. This feature is particularly essential for working with \lidar{} data, where point clouds are sparser and larger, making precise clicking on individual points impractical.

\section{Limitations}
\label{sec:limitation}
\subsection{Memory Demands} \label{subsec: memory demand}
Although the 4D setup offers substantial improvements, stacking multiple scans also increases the memory demands on our model. Voxelization and projection into consistent coordinates help to alleviate memory issues by consolidating most points from the same object into the same voxels. However, the accumulation of scans still results in a higher number of voxels, and consequently, memory requirements are still high. As GPU memory capacity increases, we expect these limitations to be mitigated, allowing for the incorporation of more data.

\subsection{Runtime} \label{subsec: run time}
The Boundary Dependent (BD) click strategy in Eq.~\eqref{eq:former_clicking} is computationally expensive during both training and inference, primarily due to the frequent need for point-wise distance calculations. This operation requires significant computational resources, which results in longer processing times and higher memory usage. By switching to our proposed click simulation strategy, we effectively reduce the training time by more than 30\%, showcasing a substantial improvement in computational efficiency.

\begin{figure}[ht!]
\centering
\caption{GPU Time Spent Accessing Memory (\%)}
\includegraphics[width=1\linewidth]{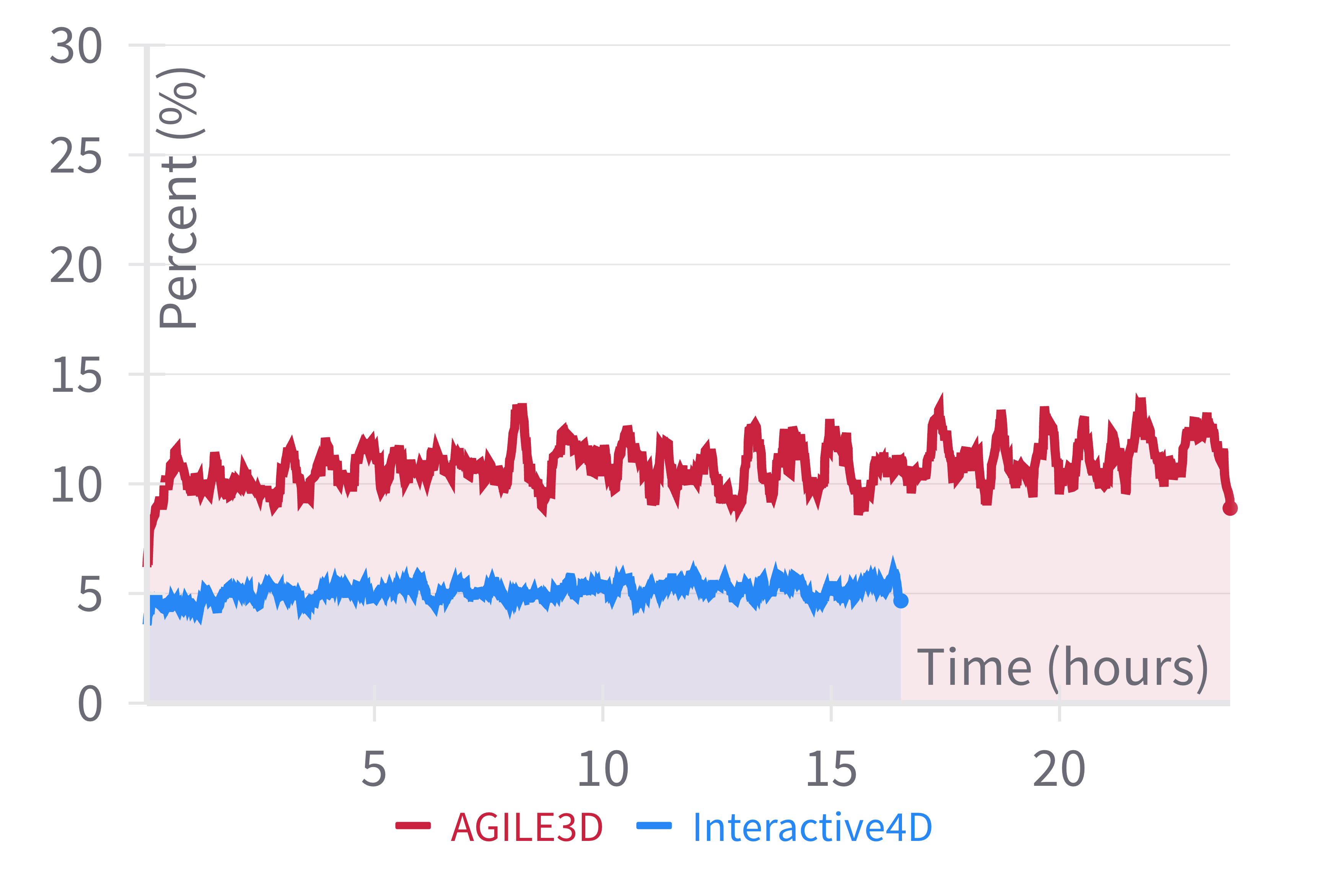}
\vspace{-0.5cm}
\label{fig:mem_time}
\end{figure}

Fig.~\ref{fig:mem_time}  further illustrates the significant reduction in GPU time spent accessing memory when using our method compared to \agilethreed{}. The figure shows that our approach consistently requires less memory access time throughout training. This reduction not only makes memory usage more efficient but also allows the training process to complete faster, as seen in the shorter overall timeline. In contrast, the higher memory access percentage for \agilethreed{} contributes to prolonged training times and inefficiency. Our method, by mitigating this overhead, achieves more streamlined computation and faster training completion.

\subsection{Tracking Failures} \label{subsec: tracking failure}
While our tracking approach performs effectively within a given temporal window, there are inherent limitations to maintaining consistent instance tracking over extended periods. Within each temporal window, instance IDs remain consistent thanks to scene annotations, which ensures stable tracking over short sequences. However, to maintain consistent instance IDs across longer sequences, we follow~\cite{aygun20214d} and implement the strategy of overlapping scans between two consecutive temporal windows. This approach allows us to associate instances between consecutive windows, specifically between the time intervals $[t, t + \tau]$ and $[t + \tau, t + 2\tau]$. In this approach, we associate instance predictions across temporal windows based on an intersection-over-union(IoU) threshold of 0.5, ensuring a one-to-one correspondence between instances. This mechanism ensures that the model can track and assign the correct instance IDs across temporal windows within the overlap.

Despite its advantages, this approach faces two key challenges that can hinder tracking consistency:

\textbf{1. Missed Instances}: If an instance ID is missed in one of the overlapping scans, for example, due to occlusion or sensor error, it becomes impossible to associate that instance with the correct ID in the next temporal window. This leads to a breakdown in tracking, as the missed instance cannot be re-associated in subsequent frames. The tracking failure occurs at this point, and the system loses the continuity of instance identification.

\textbf{2. Reappearing Instances}: Another issue arises when instances reappear in non-consecutive temporal windows. This is especially problematic in dynamic environments, such as when the ego vehicle revisits the same area after some time, encountering previously seen objects, \eg \emph{a parked car}, but at a later timestamp. Currently, there is no mechanism in place to identify such ``loop closures'' where an instance might leave the visible area, only to reappear later. This can result in the same instance being treated as a new one, thereby breaking the continuity of tracking.

Our current approach does not address these issues, which limits the ability to automatically maintain perfect tracking annotations across long sequences. This is also reflected in the tracking performance in Tab.~\ref{table:3d_4d_panoptic}, particularly the S$_\text{assoc}$ part of the LSTQ metric, which cannot be improved beyond $80\%$ $\overline{\text{IoU}}$, even with a higher click count. These failures highlight the challenges inherent in fully automating the tracking annotation process for real-world data.

Currently, in practical scenarios, human intervention is still required to handle these edge cases and maintain tracking continuity. In future work, integrating a memory-based component can more effectively manage these challenges, allowing the system to track instances across non-consecutive temporal windows and recover from missed associations. This would provide a more robust solution for long-term instance tracking in dynamic environments. 

\section{Additional Ablations} \label{sec:additional_ablation}
\subsection{Clicking Strategy using DBSCAN} \label{subsec: dbscan}
In addition to the clicking strategies discussed in Section~\ref{sec:click_sim}, another possible clicking strategy involves using DBSCAN \cite{ester1996density} (density-based algorithm for discovering clusters in large spatial databases with noise). This technique can be particularly useful in situations where objects in \lidar{} point clouds are not spatially coherent,  \eg \emph{vegetation}, may contain trees scattered throughout the scan. This can make the selection of the centroid as the ``center'' of an object inaccurate. The issue becomes even more pronounced when performing refinement clicks, where the error regions become fragmented through the scene.

To address this, we tested a two-step process after the scale-invariant selection of the largest error regions. First, we cluster the points within each error region into spatially coherent clusters using the DBSCAN algorithm. Then, we select the centroid of the largest cluster as the click within the error region. This approach prevents the potential selection of outliers when the error region is scattered.

However, while DBSCAN can offer improvements in some cases, it is highly sensitive to the parameters specified, such as the minimum number of points required for a cluster and the maximum distance between points in a cluster. Finding a single set of parameters that works well across all object categories (\eg, \textit{things} and \textit{stuff}) in the dataset can be a cumbersome and time-consuming process.

As demonstrated in Tab.~\ref{table:additional_click_simulation_ablation}, our proposed clicking strategy consistently outperforms the DBSCAN-based method. It strikes a balance between maintaining the advantages of introducing randomness to learn diverse features and ensuring robustness during testing.  Moreover, our method avoids the computational overhead associated with DBSCAN clustering.  Also, in reality, for the initial click, centroid selection  works well for \textit{things} and does not hinder performance significantly for \textit{stuff}. This allows the model to generalize effectively without the need for time-consuming DBSCAN calculations, leading to more efficient training.
\begin{table}[t]
\renewcommand{\arraystretch}{1.2}
\center
\tabcolsep=0.06cm
\footnotesize{
\caption{\small\textbf{Click Simulation with DBSCAN \cite{wang2017dbscan}}.
\textit{BD: Boundary Dependent}-Eq.~\eqref{eq:former_clicking}. \textit{SI: Scale Invariant}-Eq.~\eqref{eq: si clicking}}.
\label{table:additional_click_simulation_ablation}
\begin{tabular*}{\linewidth}{@{\extracolsep{\fill}}ccc c ccc c cc}

\toprule
\multicolumn{10}{c}{\textbf{Training}: \semantickitti{}  $\rightarrow$ \textbf{Evaluation}: \semantickitti{}} \\
\midrule
\multicolumn{3}{c}{Clicking} & & \multicolumn{3}{c}{$\overline{\text{IoU}}\text{@}\overline{\text{k}}\uparrow$} & & \multicolumn{2}{c}{$\overline{\text{NoC}}\text{@}\overline{\text{q}}\downarrow$} \\
\cmidrule{1-3} \cmidrule{5-7}\cmidrule{9-10}
\makecell{\hspace{0.5cm}SI} & \makecell{\hspace{0.5cm}Initial} & \makecell{Refinement} & & @1 & @5 & @10 & & @80 & @90 \\
\cmidrule{1-10}
\hspace{0.5cm}\cmark & \hspace{0.5cm}Centroid & Centroid & & 75.1  & 85.6 & 88.0 & & 2.88 & 4.10 \\
\hspace{0.5cm}\cmark & \hspace{0.5cm} DBSCAN & DBSCAN & & 75.2 & 85.9  & 88.3 & & 2.86 & 4.00 \\
\hspace{0.5cm}\cmark & \hspace{0.5cm} Centroid & Random & & \textbf{75.3} & \textbf{87.6} & \textbf{89.8}  & & \textbf{2.72} & \textbf{3.84} \\
\bottomrule
\end{tabular*}
}

\end{table}

\subsection{Training Click Budget} \label{subsec: click budget}
In this ablation study, we explore the impact of varying the click budget during training on the model performance. By increasing the number of clicks during training, the model has access to a greater variety of scenarios, allowing it to learn more comprehensive features and handle more diverse clicked regions, thus refining its understanding of object boundaries and error regions. This exposure leads to better generalization and improved performance during testing, even a \textbf{fixed} click budget (10 clicks) is used during testing.

However, there are trade-offs. Larger click budgets require more forward passes, increasing computational time and memory usage. As the number of clicks grows, so does the memory footprint, potentially becoming a bottleneck. For this reason, to stay consistent between different models and setups, apart from this ablation study, all the models are trained and tested with a click budget of 10 clicks per object.

Tab.~\ref{table: click_amount_ablation} shows the results for the ablation study where we increase the click budget during training while we keep the click budget fixed during testing, indeed 10 clicks per object. It shows that a larger click budget during training yields better performance, though at a higher computational cost.

\newcolumntype{g}{>{\columncolor{red!20}}c}
\begin{table}[t]
\renewcommand{\arraystretch}{1.2}
\center
\tabcolsep=0.06cm
\footnotesize{
\caption{\small\textbf{Training Click Budget Ablation.}}
\label{table: click_amount_ablation}
\begin{tabular*}{\linewidth}{@{\extracolsep{\fill}}cl c cccccc@{}}
\toprule
\multicolumn{9}{c}{\textbf{T}: \semantickitti{}  $\rightarrow$ \textbf{E}: \semantickitti{}} \\
\midrule
\multicolumn{2}{c}{} & & \multicolumn{6}{c}{$\overline{\text{IoU}}\text{@}\overline{\text{k}}\uparrow$}\\
\cmidrule(lr){4-9}
& \textbf{Clicks Per Object} & & @1 & @2 & @3 & @4 & @5 & @10\\
\midrule
& 5 clicks & & 75.7 & 82.7 & 84.7 & 85.9 & 86.7 & 90.8\\
& 10 clicks & & \textbf{77.4} & 84.9 & 87.0 & 88.3 & 89.1 & 91.2\\
& 20 clicks & & 76.9 & 85.0 & \textbf{87.5} &\textbf{ 88.9} & \textbf{89.8} & \textbf{92.2}\\
\bottomrule
\end{tabular*}
}
\end{table}

\subsection{Voxel Size Ablation} \label{subsec: voxel size}
\lidar{} point clouds are generally sparser compared to indoor scenes, necessitating careful adjustments to the voxel size to optimize data representation. The voxel size determines the resolution at which the point cloud is discretized, and selecting the appropriate voxel size is critical for balancing computational efficiency and accuracy. While smaller voxel sizes often yield better results in indoor scenes, this does not always hold for outdoor scenes. For indoor scenes, the higher point density allows for better resolution with smaller voxel sizes, which leads to improved accuracy. However, for outdoor \lidar{} point clouds, especially when dealing with sparse point clouds, reducing the voxel size further may increase computational overhead and make it harder to manage the larger number of voxels generated.

In Tab.~\ref{table:ablation voxel size}, we ablate utilizing different voxel sizes. As seen, increasing the voxel size in outdoor \lidar{} point clouds improves computational performance by reducing the number of voxels in each scene. This reduction eases memory demands and enhances running times, making the processing of large-scale point clouds more feasible. However, there is a trade-off. While increasing the voxel size improves efficiency, it can also degrade accuracy. In fact, our experiments show that the best accuracy is achieved with a voxel size of $10$~cm, then with $5$~cm. The accuracy decreases at larger voxel sizes, particularly when set to $15$~cm. This is because larger voxel sizes may lead to the loss of fine-grained spatial details, resulting in poorer object representation and reduced precision.

Thus, it is critical to select the right voxel size. Too small voxel size may introduce unnecessary computational complexity, while too large a voxel size may negatively impact accuracy. For outdoor environments, finding a balanced voxel size—such as the $10$~cm setting that offers optimal performance in our experiments—enables efficient computation while maintaining acceptable levels of accuracy.

\newcolumntype{g}{>{\columncolor{red!20}}c}
\begin{table}[t]
\renewcommand{\arraystretch}{1.2}
\center
\tabcolsep=0.06cm
\footnotesize{
\caption{\small\textbf{Voxel Size Ablation.}}
\label{table:ablation voxel size}
\begin{tabular*}{\linewidth}{@{\extracolsep{\fill}}ccccccc@{}}
\toprule
\multicolumn{7}{c}{\textbf{T}: \semantickitti{}  $\rightarrow$ \textbf{E}: \semantickitti{}} \\
\midrule
\multicolumn{1}{c}{} &  \multicolumn{6}{c}{$\overline{\text{IoU}}\text{@}\overline{\text{k}}\uparrow$}\\
\cmidrule(lr){2-7}
\textbf{Voxel Size}& @1 & @2 & @3 & @4 & @5 & @10\\
\midrule
$5$~cm & 76.9 & 84.5 & 86.6 & 87.8 & 88.6 & 90.8\\
$10$~cm & \textbf{77.4} & \textbf{84.9} & \textbf{87.0} & \textbf{88.3} & \textbf{89.1} & \textbf{91.2}\\
$15$~cm & 74.0 & 82.3 & 84.8 & 86.4 & 87.6 & 90.4\\
\bottomrule
\end{tabular*}
}
\vspace{0.35cm}
\end{table}

\subsection{Time-related Features} \label{subsec: time ablation}

In this ablation study, we examine the effect of incorporating time-related features into the \lidar{} point cloud data, with a particular focus on how different methods of encoding these features influence model performance. The results of this ablation are summarized in Tab.~\ref{table:time encoding}.

We begin by evaluating the baseline model, which does not include any time information. In this configuration, the model performs with no temporal context, serving as the reference for comparison. Next, we add time as a point feature, where each point in the point cloud is annotated with the scan from which it is originated. This time information is integrated into the backbone and subsequently incorporated into the extracted voxel features. This modification leads to noticeable performance improvements. Finally, we incorporate time information as positional encoding based on the scan index. Here, the time information is treated in a similar manner to the spatial coordinates (x, y, z)~\cite{tancik2020fourfeat}, and embedded through positional encoding. This encoded time information is then utilized in the attention blocks for both the queries and the feature representations, helping the model make more informed decisions. This method achieves the best overall performance, though the improvements are marginal compared to the previous approach.

The results demonstrate that adding time-related features enhances the model’s performance, with the greatest improvement observed when the time information is embedded through positional encoding.
\newcolumntype{g}{>{\columncolor{red!20}}c}
\begin{table}[t]
\renewcommand{\arraystretch}{1.2}
\center
\tabcolsep=0.06cm
\footnotesize{
\caption{\small\textbf{Time-related Features Ablation.}}
\label{table:time encoding}
\begin{tabular*}{\linewidth}{@{\extracolsep{\fill}}c c c cccccc@{}}
\toprule
\multicolumn{9}{c}{\textbf{T}: \semantickitti{}  $\rightarrow$ \textbf{E}: \semantickitti{}} \\
\midrule
\multicolumn{2}{c}{} & & \multicolumn{6}{c}{$\overline{\text{IoU}}\text{@}\overline{\text{k}}\uparrow$}\\

\cmidrule(lr){4-9}
 \textbf{Feature} & \textbf{Encoding} & & @1 & @2 & @3 & @4 & @5 & @10\\
\midrule

\hspace{0.5cm}\xmark& \hspace{0.5cm}\xmark & & 76.8 & 82.6 & 84.1 & 85.2 & 86.0 & 87.8\\
\hspace{0.5cm}\cmark& \hspace{0.5cm}\xmark & & 79.9 & 84.7 & 85.9 & 86.8 & 87.6 & 89.3\\
\hspace{0.5cm}\cmark& \hspace{0.5cm}\cmark & & \textbf{80.0} & \textbf{84.7} & \textbf{86.0} & \textbf{86.9} & \textbf{87.7} & \textbf{89.4}\\

\bottomrule
\end{tabular*}
}
\end{table}

\section{Additional Results} \label{subsec: additional results}
\subsection{Class-wise Results} \label{subsec: class wise results}
Tables~\ref{table:semantickitti_objects} and~\ref{table:nuscenes_objects} present class-wise results at 10 clicks, reflecting the final performance after the entire click budget is exhausted. These results are reported for all three setups, single-object, multi-object, and 4D.

Tab.~\ref{table:semantickitti_objects} displays the object-wise results for the in-distribution evaluation. The findings indicate that \ours{} consistently outperforms \agilethreed{} across all categories. While performing well on \textit{things} categories, the single-object setup struggles to maintain balance and exhibits significantly poorer performance on \text{stuff} objects. In contrast, both the multi-object and 4D setups demonstrate superior capabilities in handling this imbalance, achieving more consistent and balanced results across both \textit{things} and \textit{stuff}.

Tab.~\ref{table:nuscenes_objects} presents the results for the zero-shot evaluation, which introduces a more challenging scenario. Once again, \ours{} outperforms \agilethreed{} across all setups. In the single-object setup, the focus on individual objects allows the model to simplify the task, resulting in relatively better performance on \textit{things} compared to the multi-object or 4D setups, as it allows to handle those masks in a more local manner. However, the gap narrows with the 4D setup, as it effectively utilizes the click budget to improve overall performance. Notably, the single-object setup fails to perform well on \textit{stuff} categories, where the segmentation performance is significantly worse. The multi-object setup, on the other hand, mitigates this issue by providing the model with more context, allowing it to better delineate object boundaries. This improvement demonstrates the advantage of multi-object segmentation, where clicks on other objects can act as negative samples for refining the segmentation of a specific object. The 4D setup provides the most balanced and robust performance in the zero-shot scenario, successfully segmenting both \textit{things} and \textit{stuff}. This indicates that the temporal information incorporated in the 4D setup enables more efficient use of the click budget, allowing for better overall segmentation results in both familiar and unseen object categories.

\begin{figure}[ht!]
\centering
\caption{Object counts and appearances in the evaluated datasets.}
\begin{subfigure}[b]{0.47\textwidth}
    \centering
    \caption{SemanticKITTI - Number of points per class}
    \includegraphics[width=\textwidth]{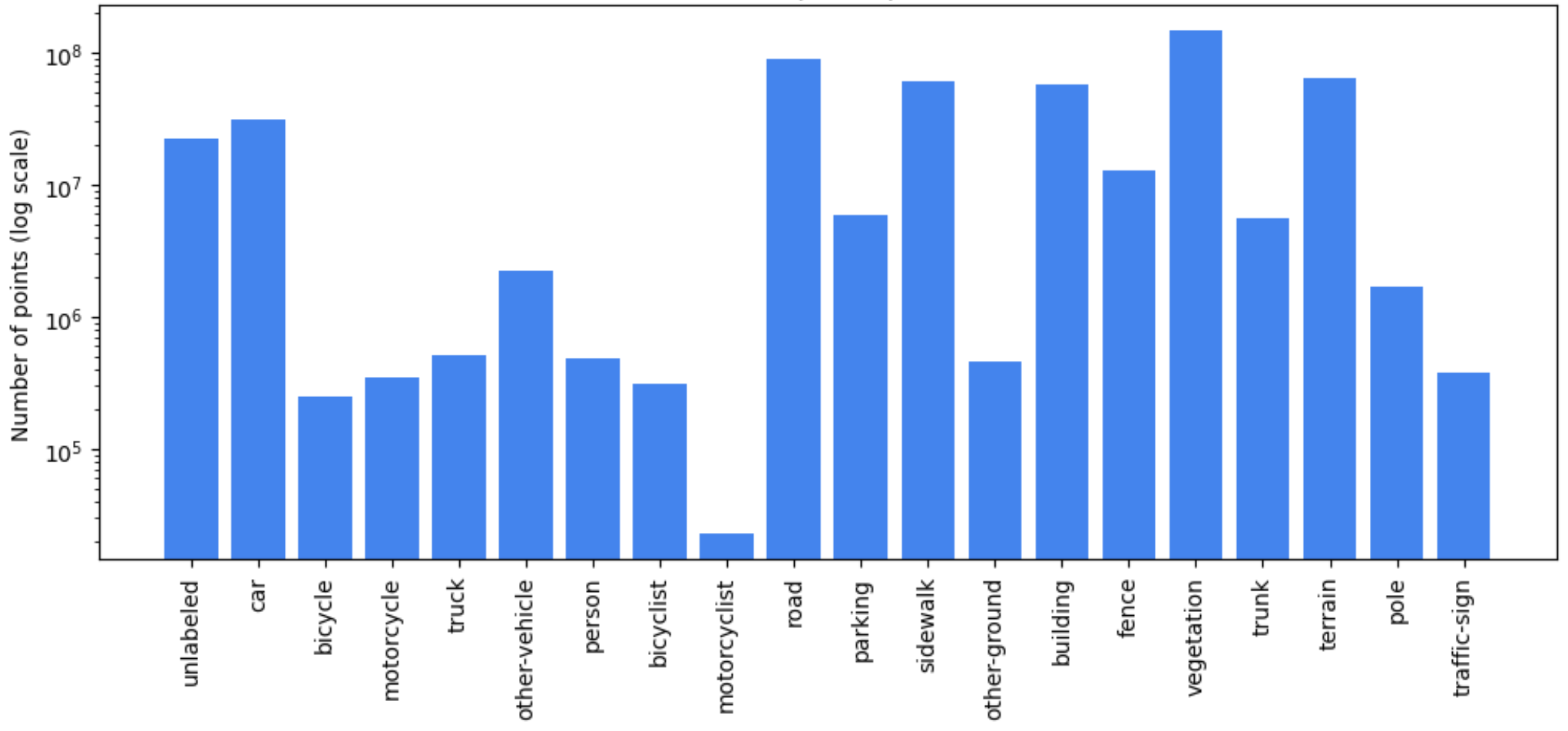}
    \label{fig:subfig1}
\end{subfigure}
\hfill
\begin{subfigure}[b]{0.47\textwidth}
    \centering
    \caption{SemanticKITTI - Number of appearances per class}
    \includegraphics[width=\textwidth]{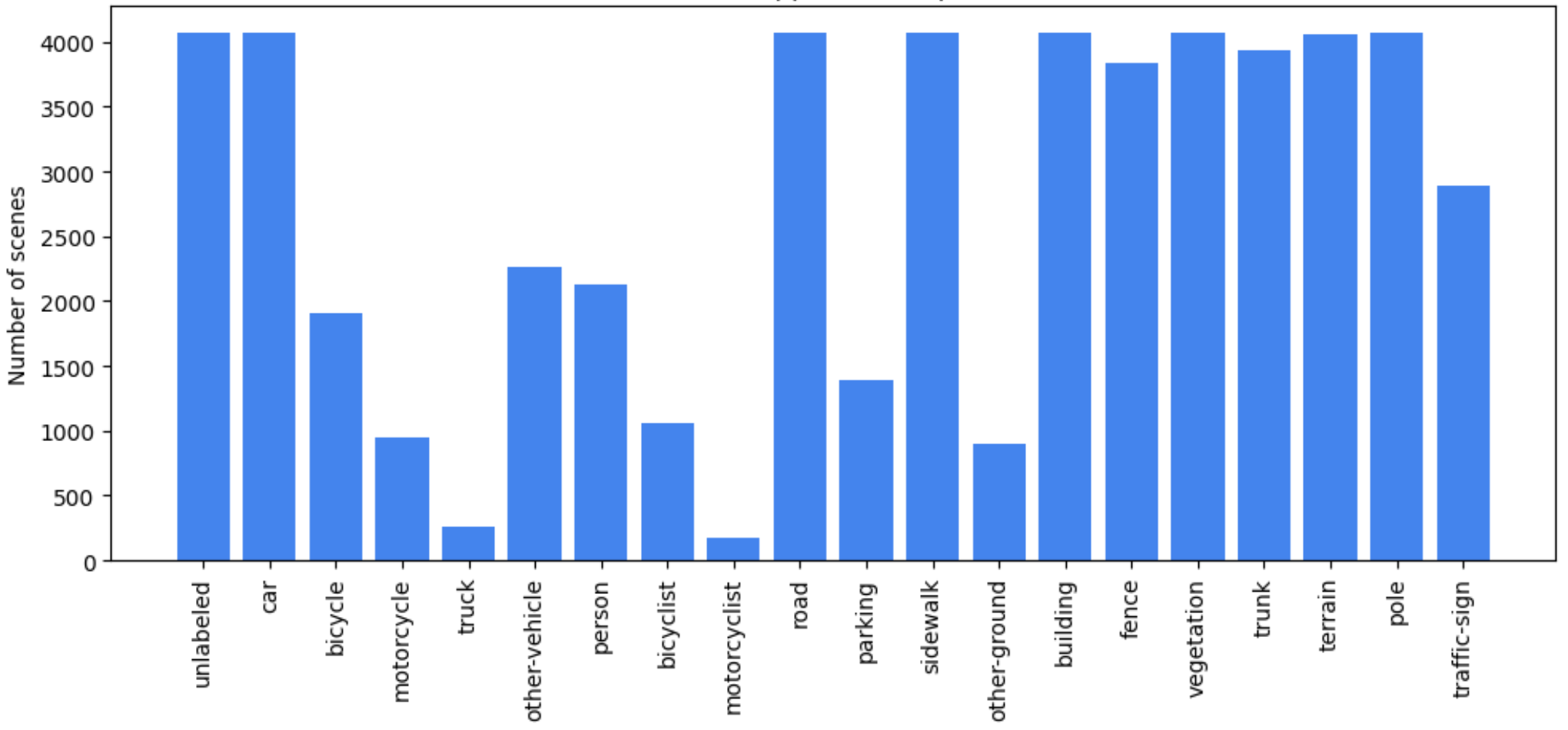}
    \label{fig:subfig2}
\end{subfigure}

\vspace{0.1cm}

\begin{subfigure}[b]{0.47\textwidth}
    \centering
    \caption{nuScenes - Number of points per class}
    \includegraphics[width=\textwidth]{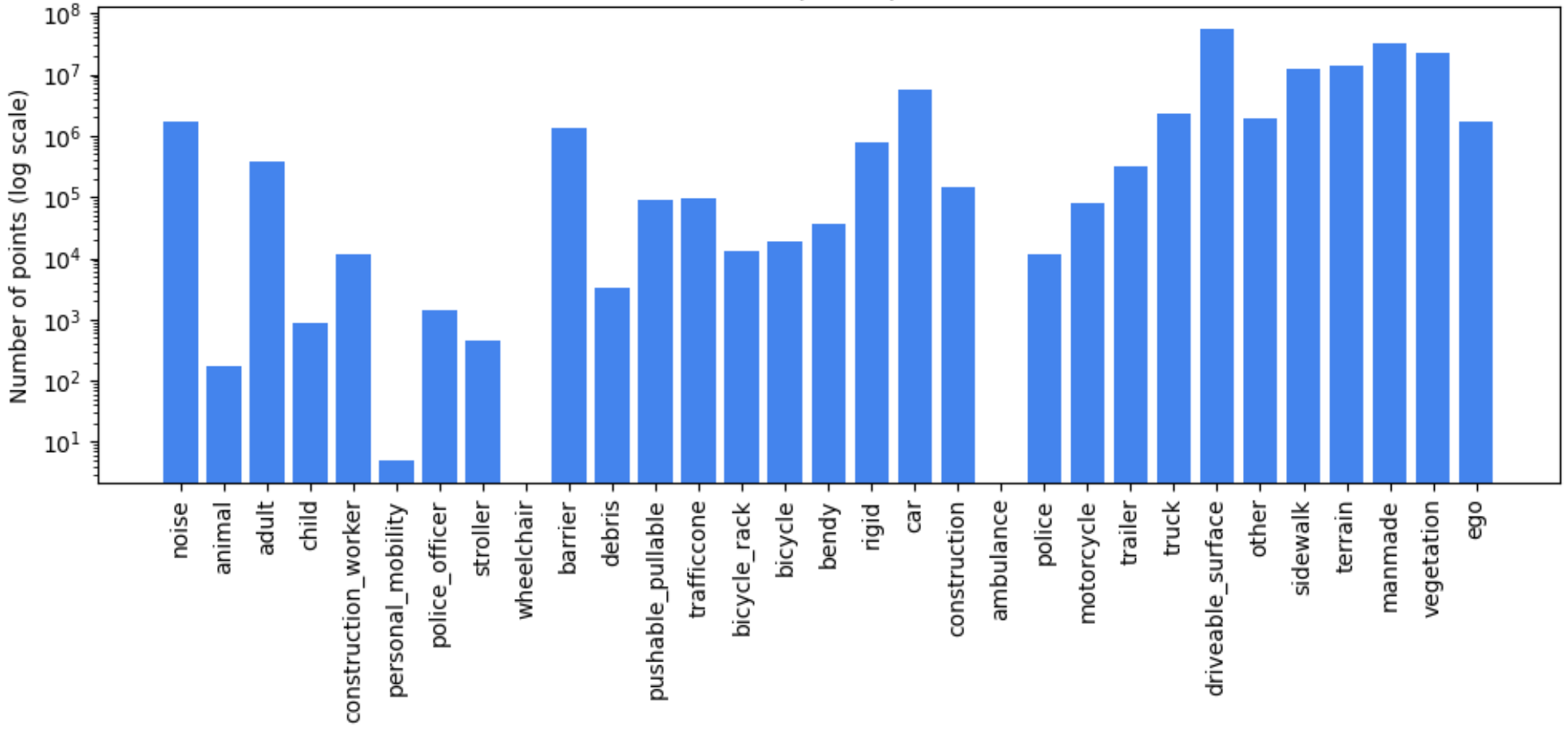}
    \label{fig:subfig3}
\end{subfigure}
\hfill
\begin{subfigure}[b]{0.47\textwidth}
    \centering
    \caption{nuScenes - Number of appearances per class}
    \includegraphics[width=\textwidth]{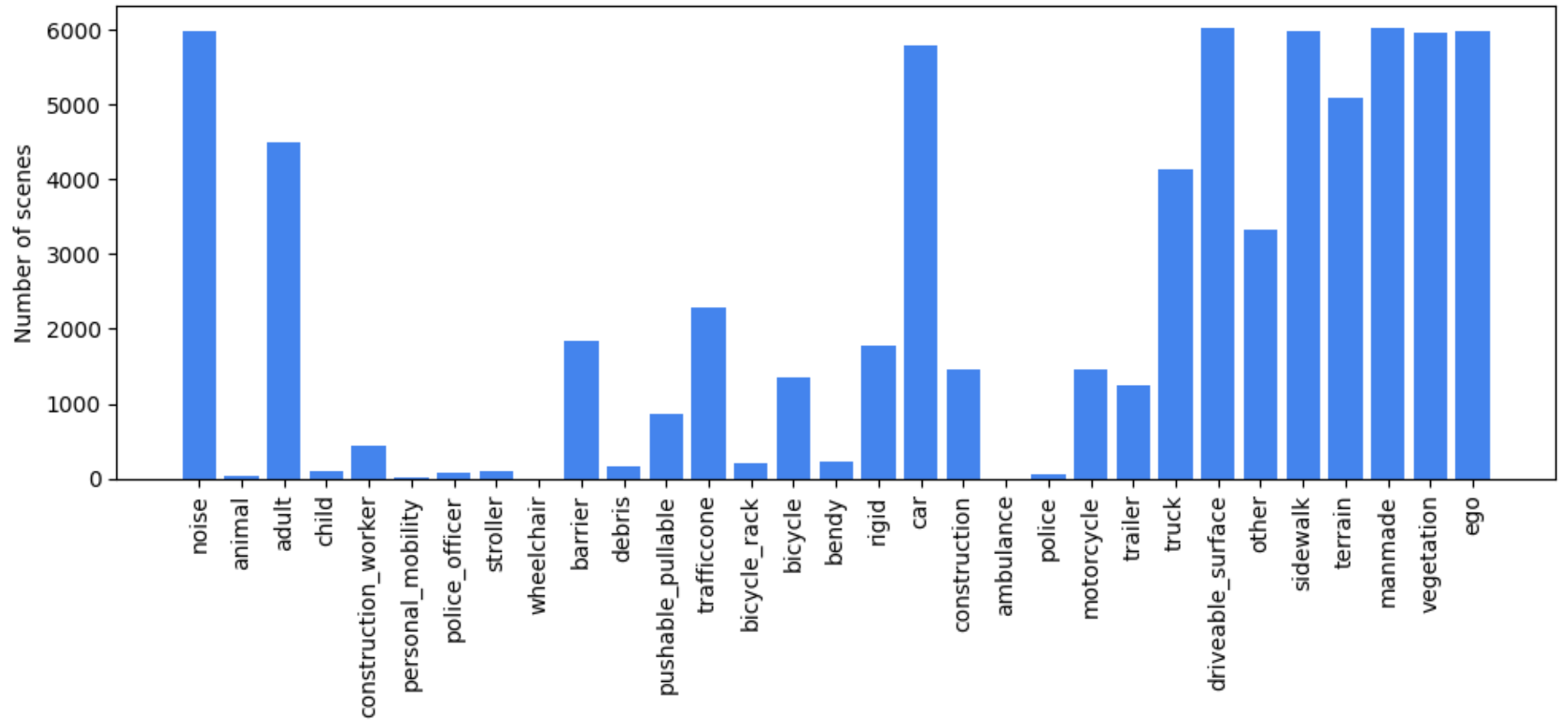}
    \label{fig:subfig4}
\end{subfigure}
\label{fig:objects_count}
\end{figure}

Additionally, we present the distribution of objects in the evaluated datasets in Fig.~\ref{fig:objects_count} to provide a better interpretation and comprehension of the results Tables~\ref{table:semantickitti_objects} and~\ref{table:nuscenes_objects}.

\subsection{Additional Qualitative Results} \label{subsec: additional qualitative results}
\begin{figure}[ht!]
\centering
\caption{Additional qualitative results of \ours{} on \semantickitti{} for the 4D setup.}
\begin{subfigure}[b]{0.925\textwidth}
    \centering
    \includegraphics[width=\textwidth]{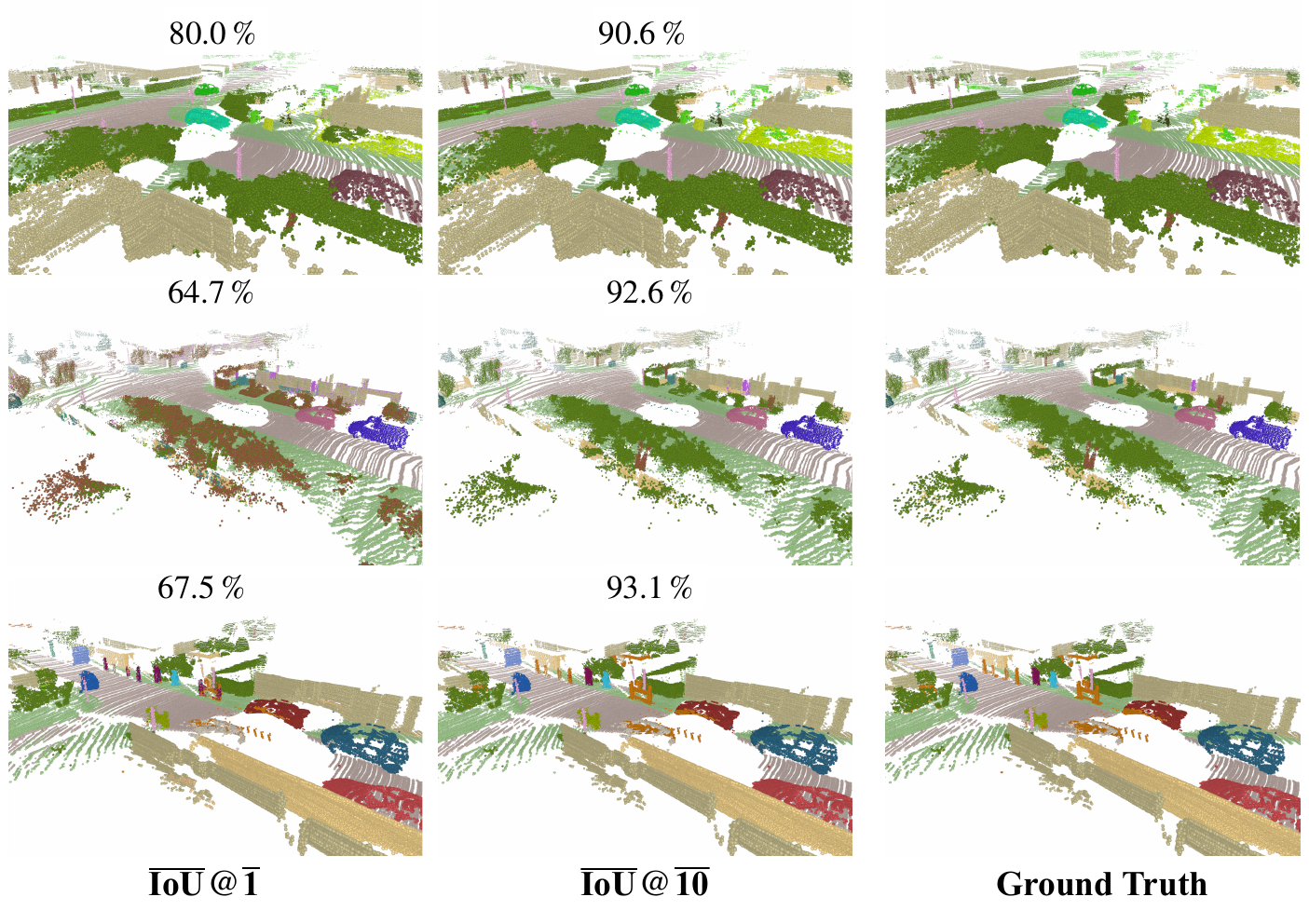}
\end{subfigure}
\label{fig:additional_qualitative}
\end{figure}

Fig.~\ref{fig:additional_qualitative} presents the additional qualitative results of \ours{} on \semantickitti{} for the 4D setup. The examples illustrate the cases where the initial segmentations are already accurate and are further improved with additional interactions. Conversely, it also highlights the instances (the second and third column) where the initial confusion arises, \eg \emph{vegetation, fences, or even moving persons}, later the segmentation is corrected with the additional clicks, ultimately achieving high performance.

\onecolumn

\twocolumn

\end{document}